\documentclass{article}

\PassOptionsToPackage{table}{xcolor}

\usepackage[main, preprint]{mystyle}

\usepackage[utf8]{inputenc} 
\usepackage[T1]{fontenc}    
\usepackage{natbib}
\usepackage{url}            
\usepackage{booktabs}       
\usepackage{amsfonts}       
\usepackage{nicefrac}       
\usepackage{microtype}      
\usepackage{xcolor}         
\usepackage{subcaption}
\usepackage{setspace}
\usepackage{graphicx}
\usepackage{multirow}
\usepackage{amsmath}
\usepackage{adjustbox}
\usepackage{caption}
\usepackage{enumitem}
\usepackage{makecell}
\usepackage{colortbl}
\usepackage{wrapfig}
\usepackage{pifont}
\usepackage[most]{tcolorbox}
\usepackage{algorithm}          
\usepackage{listings}           
\usepackage{marvosym}
\setlist[itemize]{leftmargin=*}
\definecolor{mydarkblue}{rgb}{0,0.08,0.45}

\usepackage{amsmath,amsfonts,bm}

\def\eqref#1{equation~\ref{#1}}

\def\1{\bm{1}}

\DeclareMathAlphabet{\mathsfit}{\encodingdefault}{\sfdefault}{m}{sl}
\SetMathAlphabet{\mathsfit}{bold}{\encodingdefault}{\sfdefault}{bx}{n}

\usepackage{graphicx}
\usepackage{booktabs}
\usepackage{amssymb}
\usepackage{stmaryrd}
\usepackage[table]{xcolor}
\usepackage{pifont}
\usepackage{pgfplots}
\pgfplotsset{compat=1.18}

\usepackage{colortbl}
\usepackage{tabularht}
\usepackage{array}
\usepackage{tabularx}
\usepackage{subcaption}

\definecolor{mygray}{HTML}{808080}
\definecolor{mydarkgreen}{HTML}{1B7F3A}

\usepackage{listings}
\usepackage{graphicx}
\usepackage{xcolor}
\usepackage{colortbl}
\usepackage{tabularht}
\usepackage{tabularx}
\usepackage{enumitem}
\usepackage{titletoc}

\usepackage{amsmath,amssymb,mathtools}
\usepackage{amsthm}
\usepackage{enumitem}
\usepackage{booktabs}
\usepackage{algpseudocode} 

\usepackage{array}
\newcolumntype{g}{>{\columncolor{gray!10}}c}
\usepackage{multirow}           
\usepackage{adjustbox}
\usepackage{hyperref}
\usepackage{url}
\usepackage{subcaption}
\usepackage{booktabs} 
\usepackage{array}
\usepackage{algorithm}
\usepackage{amsmath}
\definecolor{step1color}{HTML}{D5E8D4} 
\definecolor{step2color}{HTML}{A8D08D} 
\definecolor{step3color}{HTML}{82B366} 
\definecolor{step4color}{HTML}{5A8A42} 
\definecolor{promptcolor}{HTML}{F5F5F5} 
\definecolor{cond}{HTML}{2E75B6}
\definecolor{hdpocolor}{HTML}{2E75B6}
\definecolor{metiscolor}{HTML}{F3B000}

\title{%
  \raisebox{-0.30em}{%
    \includegraphics[height=1.3em]{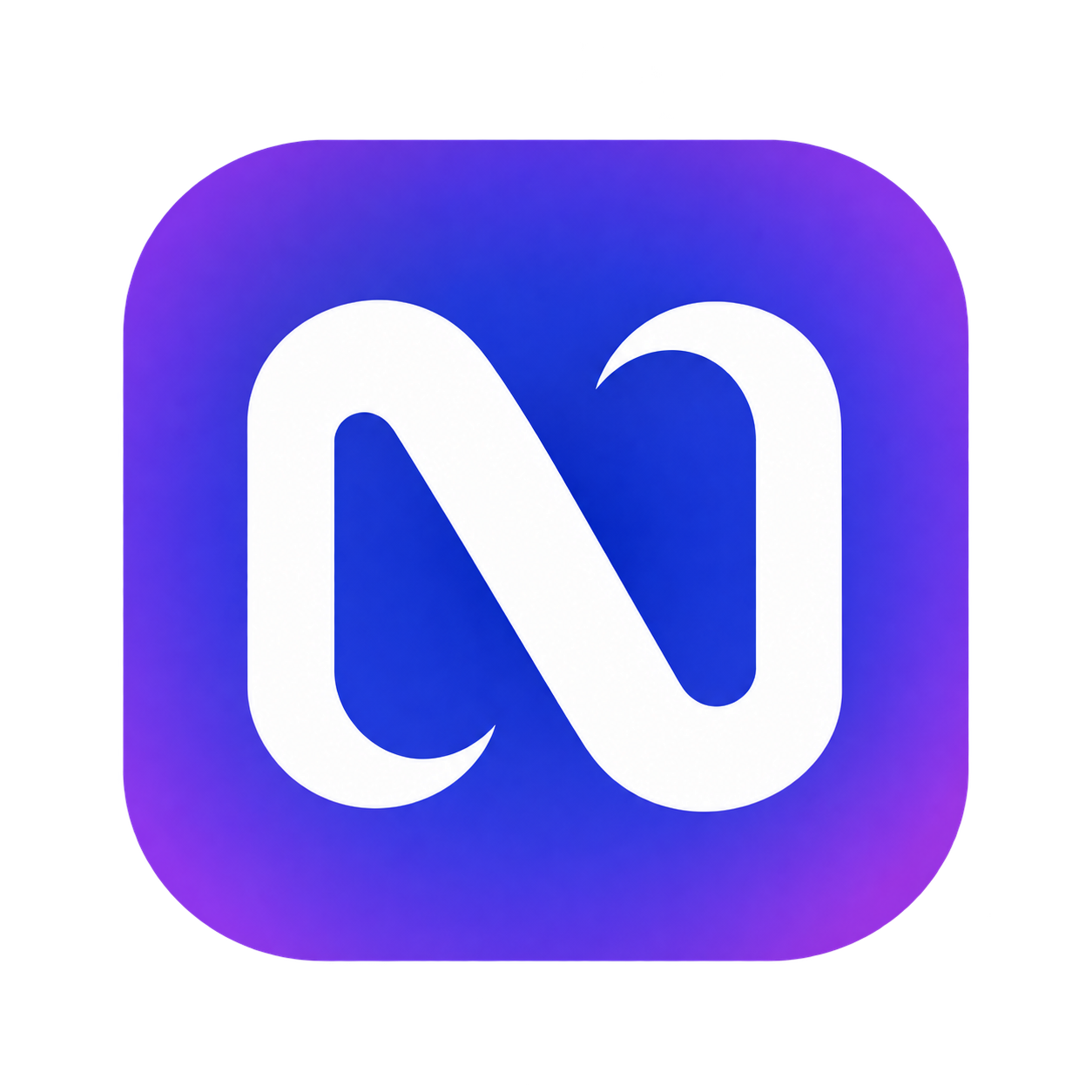}%
  }\hspace{0.1em}%
  Neos: Learning to Interleave Search
  and\\
  Visual Reasoning
}

\author{
\textbf{Ekko He}\textsuperscript{1*},
\textbf{Jack Hong}\textsuperscript{3,*,\textdaggerdbl},
\textbf{Shilin Yan}\textsuperscript{1,*,\textdaggerdbl},
\textbf{Xintong Zhang}\textsuperscript{1},
\textbf{Yujie Zhang}\textsuperscript{2},\\
\textbf{Wenjie Li}\textsuperscript{2},
\textbf{Haoran Sun}\textsuperscript{3},
\textbf{Kaixun Jiang}\textsuperscript{3},
\textbf{Jiyuan Fu}\textsuperscript{3},
\textbf{Guannan Zhang}\textsuperscript{1},
\textbf{Hongwei Xue}\textsuperscript{1,\textdagger}\\[4pt]
\normalfont
\textsuperscript{1}Accio Team, Alibaba Group\\
\textsuperscript{2}Shanghai Innovation Institute,
\textsuperscript{3}Fudan University\\
}

\headerlogos{
	\includegraphics[height=0.5cm]{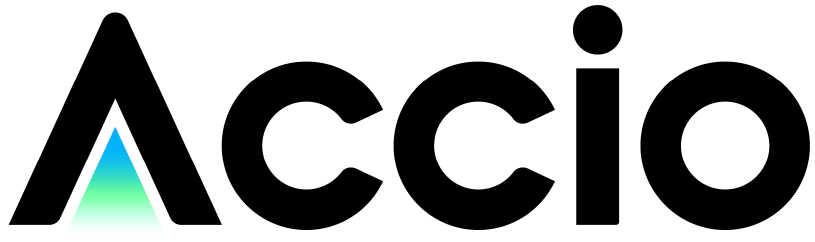}  
}

\newcommand{\github}{\raisebox{-1.5pt}{\includegraphics[height=1.05em]{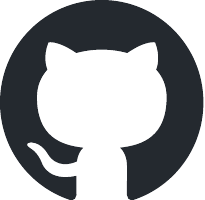}}}
\newcommand{\projectpageicon}{\raisebox{-0.12em}{\includegraphics[height=1em]{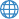}}}
\begin{document}
	
\maketitle

\insert\footins{\noindent\footnotesize $^*$Equal contribution. $^\ddagger$Project leaders. $^\dagger$Corresponding author. This work was done during Ekko He's internship at the Accio Team, Alibaba Group.}

\vspace{-0.5cm}

\begin{abstract}

Recent multimodal deep search agents have advanced tool use and visual content retrieval, yet they still struggle to convert intermediate visual evidence into actionable guidance for subsequent search, limiting their effectiveness in long-horizon settings.
To address this gap, we present \textbf{Neos}, a multimodal deep search framework that learns both \emph{how to acquire evidence} and \emph{how to use it to guide subsequent decisions}.
To make this coupling learnable, Neos constructs training tasks with intermediate visual dependencies while filtering out textual shortcuts, and introduces a lightweight, trainable Multimodal Evidence Interface (MEI) that transforms retrieved content into goal-conditioned observations.
At its core, an \emph{agent--environment co-evolution} procedure treats MEI as a co-adaptive component of this loop.
Specifically, reference-guided credit assignment uses verified successful trajectories to localize critical policy errors while preserving credit for useful decisions in failed searches, whereas rollout feedback selects informative local evidence-processing interactions for MEI supervised fine-tuning (SFT).
By alternating policy reinforcement learning (RL) with MEI SFT, Neos allows the policy's evolving evidence requests to shape MEI training and the adapted MEI to reshape subsequent observations, closing the loop between evidence acquisition and evidence-conditioned action.
Across ten benchmarks, Neos-9B and Neos-35B demonstrate leading performance among specialized multimodal deep-search agents.\\

    \projectpageicon{}  \textbf{Project Page:} \href{https://Accio-Lab.github.io/Neos}{https://Accio-Lab.github.io/Neos} \vspace{0.1cm}

    \github{}  \textbf{GitHub Repo:} \href{https://github.com/Accio-Lab/Neos}{https://github.com/Accio-Lab/Neos} 	

\end{abstract}

\begin{figure}[h]
    \centering
    \includegraphics[width=1\linewidth]{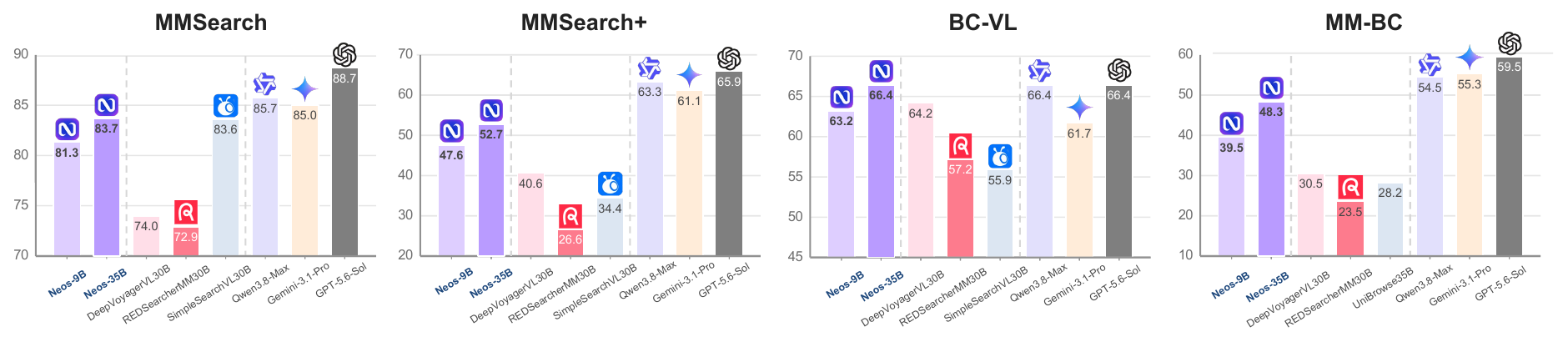}
    \caption{\textbf{Benchmark performance of Neos-9B and Neos-35B.}}
    \label{fig:benchmark_chart}
\end{figure}

\clearpage

\section{Introduction}
\label{sec:introduction}

Real-world information-seeking problems are often open-ended and grounded in both textual and visual evidence.
Deep search agents have made substantial progress in text-centric settings by iteratively searching the web, browsing sources, and synthesizing grounded answers~\citep{jin2025search,zheng2025deepresearcher,wu2026webdancer,li2025websailor,team2025tongyi,team2025mirothinker,li2026webthinker}.
Recent multimodal systems extend these agents with multimodal tool use, multi-turn visual interaction, scalable data synthesis, and visual context management~\citep{hong2026deepeyesv2,chng2025sensenova,zhang2026vsearcher,zhang2026visual,yao2026mm,huang2026towards,liu2026points,du2026towards}.
Despite these advances, many systems fail to consistently leverage visual evidence during search. Yet such evidence is pivotal for guiding subsequent queries, interpreting retrievals, and determining the need for further inspection~\citep{li2025mm,tao2026mmsearch,zeng2026vision}.
The central challenge is therefore to sustain an \emph{evidence--decision loop}, in which search decisions determine which evidence is acquired, and newly acquired evidence guides subsequent decisions.

Existing methods still exhibit three intertwined limitations:
\textbf{I) Weak intermediate visual dependencies.}
Training and evaluation tasks often do not require newly acquired visual evidence to influence later search, while synthesized tasks may confine vision to initial recognition or final attribute extraction~\citep{wu2025mmsearch,zhang2026visual}, while captions, metadata, or indexed text can bypass intermediate visual reasoning~\citep{zeng2026vision}.
Consequently, agents can succeed without internalizing the paradigm of using newly acquired visual evidence to guide subsequent decisions.
\textbf{II) Ambiguous process credit.}
Prior work provides fine-grained learning signals through advantage estimation, reward shaping, and process-aware optimization~\citep{feng2026group,xia2026search,lu2026abseeker,liu2026cafe,yan2026act}.
However, these signals often lack an evidence-grounded reference for attributing trajectory-level outcomes to individual decisions.
Final-answer supervision therefore remains too global to identify critical errors, distinguish them from downstream consequences, or preserve credit for useful decisions in failed searches.
\textbf{III) Decoupled policy and evidence interface.}
Retrieved content is commonly converted into observations by fixed, often proprietary auxiliary models~\citep{zhang2026deepvoyager}.
Beyond their inference cost, these interfaces cannot adapt to the policy's evolving evidence needs, preventing evidence presentation and search behavior from improving together.

To address these limitations, we present \textbf{Neos}, a multimodal deep search framework that learns both how to acquire evidence and how to use it to guide subsequent decisions.
Neos closes this evidence--decision loop through three complementary components:
I) Evidence-dependent data synthesis.
We construct tasks with intermediate visual dependencies and screen for textual shortcuts, producing questions and trajectories in which newly acquired visual evidence is necessary for later search.
II) Reference-guided credit assignment.
Using verified successful trajectories and the evidence available at each turn, our credit-assignment method localizes critical errors in evidence seeking and use, attenuates penalties for their downstream consequences, and preserves positive credit for useful decisions in failed searches.
This provides targeted reinforcement-learning signals for both evidence acquisition and evidence-conditioned decision making.
III) Agent--environment co-evolution.
We embed a lightweight, trainable Multimodal Evidence Interface (MEI) into the search tools to transform retrieved content into goal-conditioned observations, while leaving search and on-demand visual inspection under policy control.
After supervised initialization, Neos alternates policy reinforcement learning (RL) with MEI supervised fine-tuning (SFT).
Rollout outcomes and shared process feedback select local evidence-processing interactions for MEI training, including useful interactions from failed searches.
The policy's evolving evidence needs thereby shape MEI training, while the adapted MEI reshapes observations for policy learning. Our contributions are fourfold:

\begin{itemize}[topsep=3pt, itemsep=1pt]
    \item We introduce an \textbf{evidence-dependent synthesis pipeline} that produces multimodal search tasks in which newly acquired visual evidence is necessary for subsequent search decisions.

    \item We propose \textbf{reference-guided credit assignment}, which localizes critical policy errors and their downstream consequences, while preserving positive credit for productive and corrective decisions in failed searches.

    \item We develop a lightweight \textbf{Multimodal Evidence Interface (MEI)} and an \textbf{agent--environment co-evolution} procedure, allowing joint improvement of evidence presentation and the search policy.

    \item Across ten benchmarks, Neos-9B and Neos-35B outperform their respective base workflows by 17.4 and 15.5 points on average, achieving leading open-source multimodal deep-search performance.
\end{itemize}

\section{Related Work}
\label{sec:related_work}

\textbf{From multimodal understanding to active information seeking.}
MLLMs have advanced visual understanding and reasoning but remain limited by parametric knowledge and inference-time evidence.
AVIS and Vision Search Assistant extend visual question answering with tools and web search~\citep{hu2023avis,zhang2024vision}.
Recent ``thinking with images'' methods integrate visual interaction into reasoning: DeepEyes learns image manipulation via RL~\citep{zheng2026deepeyes}; Adaptive-CoF performs question-guided region selection and zooming, and MGPO learns multi-turn visual grounding from answer-level rewards~\citep{zhang2025chain,huang2026mgpo}; RSAgent uses visual feedback for segmentation~\citep{he2025rsagent}; PyVision and Thyme use generated Python tools and executable code for visual computation~\citep{zhao2025pyvision,zhang2025thyme}; Mini-o3 extends exploratory visual reasoning to longer interactions~\citep{lai2026mini}; and MAPO aligns tool outputs with visual observations in policy optimization~\citep{yang2026walk}.
These methods make visual inspection and manipulation part of reasoning, but multimodal deep search must iteratively acquire and verify external evidence to guide subsequent search.

\textbf{Multimodal deep search agents.}
Multimodal agents learn multi-turn evidence acquisition through RL, e.g., MMSearch-R1 and WebWatcher~\citep{wu2025mmsearch,geng2026webwatcher}.
ProMMSearchAgent uses process-oriented rewards, and OpenSearch-VL preserves useful prefixes via fatal-aware masking and advantage clamping~\citep{yan2026prommsearchagent,chen2026opensearch}.
Scalable supervision includes graph-based task construction and trajectory synthesis~\citep{chu2026redsearcher,yao2026mm}; VSearcher combines iterative-injection data synthesis with SFT-then-RL~\citep{zhang2026vsearcher}, while ODE adapts data generation from rollout feedback~\citep{huang2026towards}.
Evidence handling also differs: Vision-DeepResearch supports multi-turn, multi-entity, and multi-scale search~\citep{huang2026vision}; LMM-Searcher separates image references from on-demand acquisition~\citep{du2026towards}; POINTS-Seeker renders older observations as images while keeping recent text~\citep{liu2026points}; DeepVoyager-VL constructs intermediate visual dependencies with helper-mediated observations and reusable visual state~\citep{zhang2026deepvoyager}; and SimpleSearch-VL places webpage summarization and thumbnail verification within the policy~\citep{dai2026simplesearch}.

\section{Method}
\label{sec:method}

\subsection{Problem Setting and Agent Workflow}
\label{sec:overview}

We formulate multimodal deep search as an \emph{evidence--decision loop}. Given a question $q$ and initial images $\mathcal I_0$, the policy $\pi_\theta$ interacts with a tool environment $\mathcal E_\phi$ to produce an answer. At turn $t$, history $h_t$ contains the inputs and preceding action--observation pairs,
\begin{equation}
\label{eq:agent_environment_interaction}
a_t \sim \pi_\theta(\cdot \mid h_t),
\qquad
o_t \sim \mathcal E_\phi(\cdot \mid h_t,a_t),
\end{equation}
with $h_{t+1}=h_t\oplus(a_t,o_t)$. The interaction ends when the policy produces a final answer.

Neos discovers candidate evidence through \texttt{TextSearch}, \texttt{ImageSearch}, and \texttt{ReverseImageSearch}, and accesses source pages via \texttt{WebVisit}. These tools return text and image URLs, while \texttt{FetchImage} and \texttt{CropImage} expose selected images and regions on demand~\citep{du2026towards}. Loaded images and crops remain accessible through reusable identifiers in an episode-local visual register~\citep{zhang2026deepvoyager}. To present retrieved multimodal content as policy observations, Neos incorporates a trainable Multimodal Evidence Interface (MEI)
within \texttt{WebVisit}, \texttt{ImageSearch}, and
\texttt{ReverseImageSearch} to convert search results into goal-conditioned observations.
The parameters $\phi$ govern this interface within $\mathcal E_\phi$,
while the underlying web and search services remain fixed
(Section~\ref{sec:mei}).

\subsection{Multimodal Evidence Interface}
\label{sec:mei}

Online content becomes useful for decision-making only after the
agent identifies evidence relevant to its current search goal.
Existing search pipelines often delegate this processing to large
or proprietary auxiliary models~\citep{wu2025mmsearch,zhang2026deepvoyager},
incurring repeated inference latency and serving cost.
When these models remain fixed, evidence processing cannot adapt
to the evolving policy's needs.
MEI addresses these limitations through a lightweight, trainable
vision--language model, $g_\phi$, that adapts evidence presentation
to the policy's current search objective.

Embedded in \texttt{WebVisit}, \texttt{ImageSearch}, and \texttt{ReverseImageSearch}, MEI is implemented by a lightweight vision--language model.
Conditioned on the policy's local goal and the returned material, it extracts relevant page evidence, filters and ranks image candidates, and assesses visual matches.
Thus, evidence is selected and summarized for the current search objective rather than presented as an undifferentiated collection of retrieved material. It inspects candidate images internally, but returns structured text and image references as ordinary tool observations. The agent does not invoke MEI as a separate model and retains control over query formulation and on-demand visual inspection. This separates learning what evidence to present from learning how to act on it.

\begin{figure}[t]
    \centering
    \includegraphics[width=0.95\linewidth]{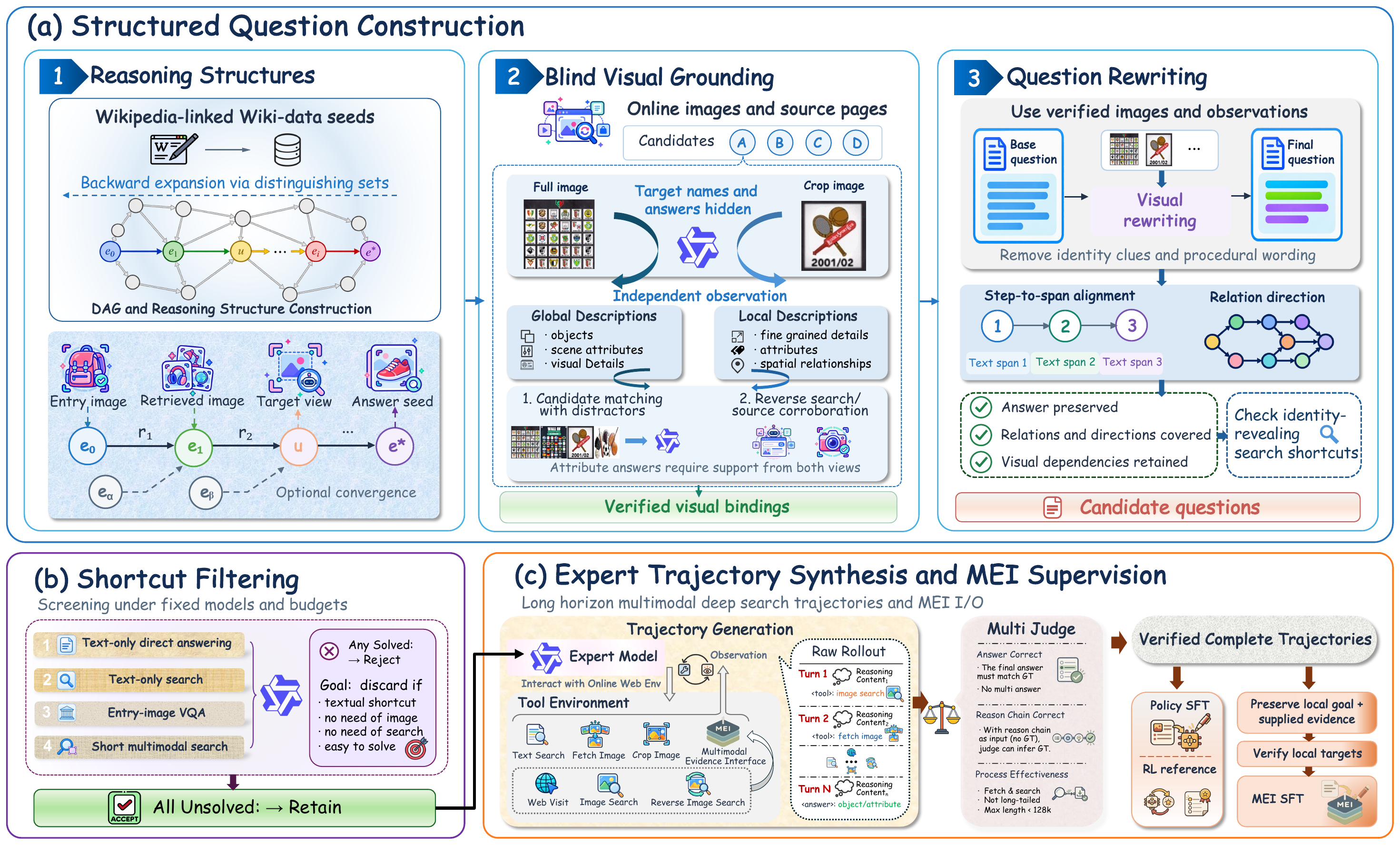}
    \caption{
    \textbf{Data synthesis pipeline.}
    Reasoning structures with visual checkpoints undergo blind visual grounding, question rewriting, and shortcut screening.
    Verified successful searches provide policy demonstrations, success references, and local MEI training interactions.
    }
    \label{fig:datapipeline}
    \vspace{-8pt}
\end{figure}

\subsection{Evidence-Dependent Task Synthesis}
\label{sec:data}

Our data pipeline is designed to instantiate the \emph{evidence--decision loop} at the task level: an agent must acquire intermediate visual evidence, and that evidence must constrain subsequent search and reasoning.
The presence of images alone does not guarantee this property.
Tasks may restrict vision to initial recognition or final attribute extraction, while textual clues can reveal identities intended to be inferred visually.
We therefore construct tasks in which intermediate visual evidence supplies entity bindings for later reasoning steps, and combine structural checks, blind visual grounding, and shortcut screening before collecting verified searches (Figure~\ref{fig:datapipeline}).

\textbf{Reasoning structures and identifiability.}
To make intermediate evidence decision-relevant, we construct reasoning structures in which visual checkpoints are required for later relational inference.
Starting from Wikipedia-linked Wikidata entities, we recursively expand \emph{distinguishing sets}: neighboring relational constraints that jointly identify an entity within a knowledge-graph (KG) snapshot $\mathcal K$.
The resulting DAG connects prerequisites to identified entities while preserving the direction of each KG fact.
Reasoning structures comprise paths or convergent subgraphs with a visual entry anchor and at least one intermediate visual checkpoint; converging branches have disjoint entity sets before their shared entity.
These checkpoints connect visual entity identification to subsequent relational reasoning, making newly acquired visual evidence a prerequisite for later steps.
To check that the specified evidence bindings identify an unambiguous target, we evaluate
\begin{equation}
\label{eq:data_identifiability}
\mathcal{U}_{\mathcal{K}}(C,B)
=
\left\{
u \;\middle|\;
\exists \mathbf{z},\; z_{\mathrm{out}}=u,\;
\bigwedge_{(v,r,w)\in C}(z_v,r,z_w)\in\mathcal{K},\;
\bigwedge_{v\in B}z_v=b_v
\right\}.
\end{equation}
Here, $C$ denotes relational constraints, $B$ denotes bindings at the entry and checkpoints, $\mathbf z$ assigns entities to graph variables, and $z_{\mathrm{out}}$ denotes the final target entity.
We retain structures satisfying $\mathcal U_{\mathcal K}(C,B)=\{u^\star\}$, ensuring that the seed $u^\star$ is uniquely identified within $\mathcal K$ conditional on the specified bindings.
This check establishes structural identifiability; the following stages assess whether the required evidence is visually grounded and whether textual bypasses remain.

\textbf{Blind visual grounding.}
To ensure that the required bindings are genuinely obtainable from visual evidence, an observer model describes full images and crops independently, without target identities or expected answers.
These descriptions are matched against candidates containing distractors, with the target unmarked.
Bindings require distinctive visual evidence or compatible observations corroborated by reverse search or source pages; visual attribute answers require support from both full-image and crop observations.

\textbf{Question rewriting and filtering.}
Even with evidence-dependent structures, a question can still be solvable through leaked names or text-only shortcuts.
We verbalize each reasoning structure, then use a multimodal rewriter conditioned on verified images, crops, and observations to remove revealing names and procedural wording while preserving the answer and dependencies.
An independent check aligns each step with a span in the final question to verify coverage and relation direction, while targeted text searches screen for wording that exposes hidden identities.
We then apply four sequential probes---text-only answering, text-only search, entry-image VQA, and multimodal search---and reject questions solved by any probe, with limited budgets for the search probes.
The text-only and entry-image probes screen for routes that bypass intermediate visual acquisition, while the budget-limited multimodal probe additionally filters task difficulty.

\textbf{Trajectory curation.}
For retained questions, an expert model conducts long-horizon searches, producing trajectories that follow the desired evidence--decision loop.
Successful trajectories serve as demonstrations and success references for credit assignment, while local tool interactions provide MEI training inputs.
The resulting supervision connects task construction to both search-policy learning and evidence-interface training. Further details appear in Appendix~\ref{appendix:data}.

\begin{figure}[t]
    \centering
    \includegraphics[width=1\linewidth]{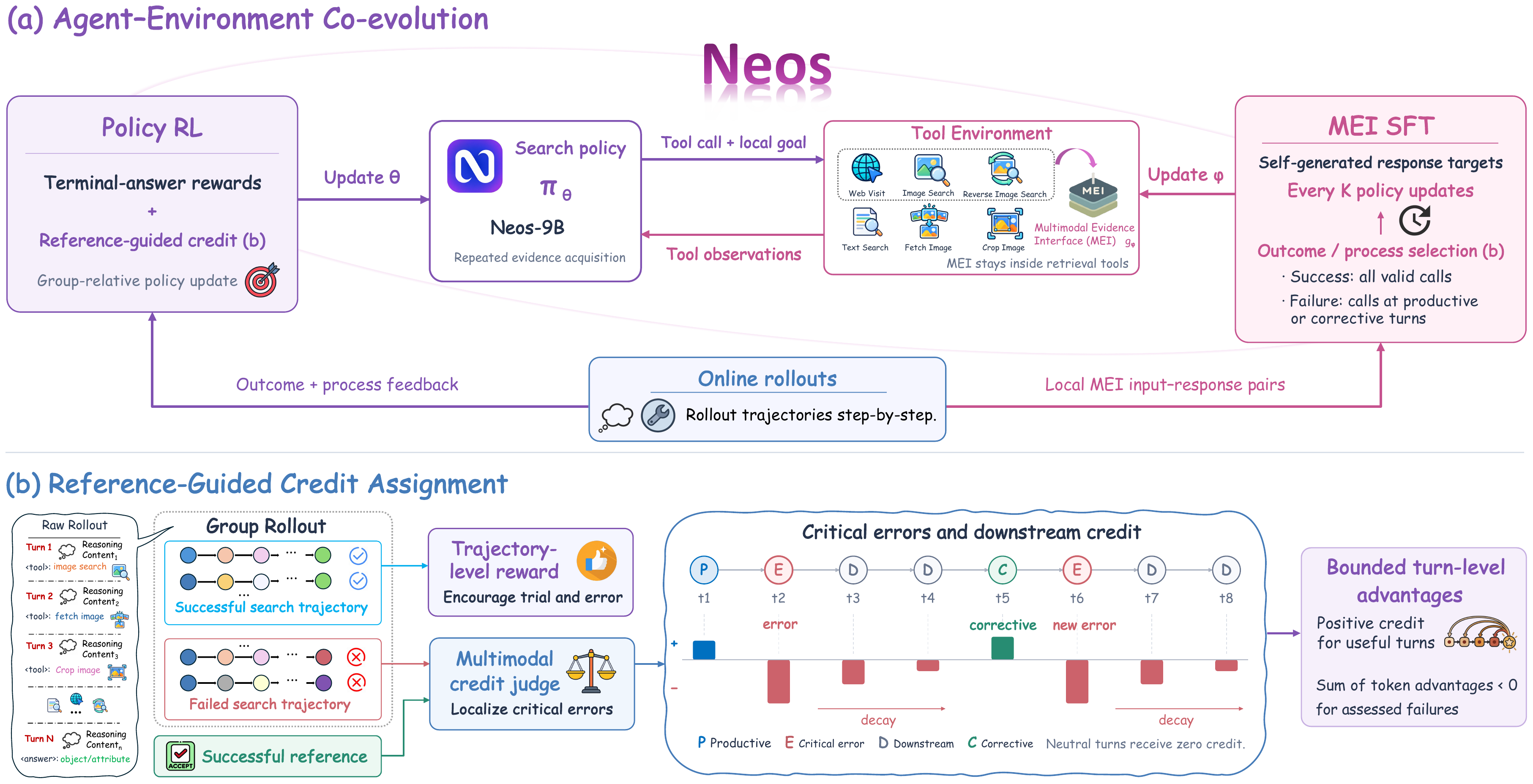}
    \caption{
    \textbf{Agent--environment co-evolution.}
    Policy RL alternates with MEI SFT on outcome- and process-filtered online interactions.
    Reference-guided credit supports policy optimization and identifies useful local interactions for MEI updates.
    }
    \label{fig:framework}
    \vspace{-10pt}
\end{figure}

\subsection{Reference-Guided Credit Assignment}
\label{sec:optimization}

An incorrect final answer does not invalidate all preceding decisions; a failed search may still acquire relevant evidence or correct earlier mistakes. Although process-oriented rewards and prefix-preserving updates provide finer-grained supervision~\citep{yan2026prommsearchagent,chen2026opensearch}, uniformly propagating trajectory-level penalties still obscures turn-level contributions. Neos instead uses verified successful trajectories as references to identify the role of each turn.

\textbf{Supervised initialization.}
The agent and MEI are initialized on their respective curated examples using a masked autoregressive objective normalized over target tokens.
Policy targets comprise reasoning, tool calls, and final answers, while MEI targets comprise structured evidence responses.
Conditioning content, including tool observations, and padding are excluded from the loss.

\textbf{Reference-guided assessment.}
For each question, we sample $G$ rollouts under $\pi_{\theta_{\mathrm{old}}}$ and $\mathcal E_\phi$, retaining complete groups containing both successes and failures.
Binary answer-correctness rewards $r_i$ define group-relative advantages
$A_i=(r_i-\bar r)/(\operatorname{std}(\mathbf r)+\varepsilon_{\mathrm{std}})$,
where $\bar r=G^{-1}\sum_j r_j$.
For failed trajectories, a credit judge assesses recorded actions and observations against a verified successful reference.
The reference grounds assessment in a demonstrated successful investigation, rather than prescribing a required sequence of actions.
The judge identifies critical policy errors, their downstream consequences, and productive or corrective decisions while respecting the evidence available at each turn.
For successful trajectories, alternative valid search routes are permitted.

\textbf{Turn-level credit.}
We focus negative credit on critical errors and attenuate penalties on their downstream consequences.
Productive and corrective decisions receive positive credit, so useful evidence acquisition and recovery can be reinforced even within failed searches, rather than merely exempted from penalties.
For an assessed failure, we set $b_i=|A_i|$ and assign
\begin{equation}
\label{eq:training_turn_credit}
w_{i,t}=\begin{cases}
-\max(\kappa_{\mathrm e},s_{i,t}c_{i,t}),
& \text{critical error},\\
-\lambda_{\mathrm d}\max(\kappa_{\mathrm d},s_{i,t}c_{i,t})\gamma^{d_{i,t}-1},
& \text{downstream consequence},\\
\lambda_{\mathrm c}c_{i,t}, & \text{corrective},\\
\lambda_{\mathrm p}c_{i,t}, & \text{productive},\\
0, & \text{neutral}.
\end{cases}
\end{equation}
Here, $s_{i,t}$ and $c_{i,t}$ denote severity and confidence.
Unannotated turns following an active error inherit its severity and confidence, with $d_{i,t}\geq1$ measuring elapsed turns; otherwise, they are neutral.
Corrective turns stop propagation, and subsequent critical errors restart it.
Each policy token $k$ receives $\widetilde A_{i,k}=b_iw_{i,t(k)}$. Additional engineering techniques of RL are provided in Appendix~\ref{appendix:rltrick}.

\textbf{Policy optimization.}
We rescale positive and negative credit into bounded advantages $\widehat A_{i,k}$, keeping their sum negative for each assessed failure while preserving the uniform advantage $A_i$ for successful trajectories.
This reinforces useful local decisions while keeping failed trajectories negative in aggregate.
Let $\mathcal M_i$ index policy-generated tokens and $\rho_{i,k}(\theta)$ denote the likelihood ratio between $\pi_\theta$ and $\pi_{\theta_{\mathrm{old}}}$ on the recorded prefix.
Using the dual-clipped surrogate $\ell_{\mathrm{dc}}$, we optimize
\begin{equation}
\label{eq:training_policy_objective}
\mathcal L_{\mathrm{RL}}(\theta)
=-\frac{\sum_i\sum_{k\in\mathcal M_i}
\ell_{\mathrm{dc}}\!\left(\rho_{i,k}(\theta),\widehat A_{i,k}\right)}
{\sum_i |\mathcal M_i|}.
\end{equation}
Outcome rewards remain unchanged, and credit estimates are held fixed during optimization. The SFT objective, credit rescaling, and dual-clipping definitions appear in Appendix~\ref{appendix:training_details}.
The same process labels also select useful local MEI interactions, connecting decision-level credit with evidence-interface adaptation in Section~\ref{sec:coevolution}.

\subsection{Agent--Environment Co-evolution}
\label{sec:coevolution}

Optimizing either component while holding the other fixed limits adaptation: a fixed policy exposes MEI to a static distribution of evidence requests, whereas a fixed MEI cannot adjust its observations to the policy's evolving needs. Neos instead co-evolves the two components, allowing evidence presentation and search behavior to adapt to each other and improve jointly (Figure~\ref{fig:framework}).

\textbf{Outcome- and process-filtered supervision.}
Since the evidence environment must adapt to the policy's evolving evidence requests, its supervision is drawn from online agent--MEI interactions rather than static demonstrations alone.
After supervised initialization, we record each agent--MEI interaction, including inputs to MEI, its generated response, the corresponding policy turn and the outcome.
Correct trajectories contribute all valid MEI calls; failed trajectories selected for agent optimization contribute only calls labeled \emph{productive} or \emph{corrective}.
This filtering allows MEI to learn from evidence-processing interactions that remain useful even when the final search fails.

\textbf{Alternating adaptation.}
We collect agent rollouts asynchronously under a fixed MEI version and update the agent after each.
After every $K$ agent updates, MEI is fine-tuned from its latest weights on newly retained calls collected since its previous update.
The evolving agent determines which evidence-processing interactions become available for MEI training, and the updated MEI changes the observations used for subsequent agent decisions.
This alternating procedure creates a reciprocal adaptation loop: policy updates reshape MEI's training signal, and MEI updates reshape the policy's decision context, while the underlying web and tool APIs remain unchanged.

\begin{table*}[t]
\centering
\caption{
Comparison of multimodal search methods across 10 benchmarks. The best and second-best results are \textbf{bolded} and \underline{underlined}. Green $\Delta$ rows report absolute score improvements over Qwen3.5-9B and Qwen3.5-35B-A3B under our workflow for the SFT and RL models.
}
\label{tab:main_results}

\begingroup
\setlength{\tabcolsep}{3pt}
\renewcommand{\arraystretch}{1.12}
\newcommand{\mmdgain}[1]{\textcolor{green!50!black}{\fontsize{6.4}{6.6}\selectfont\textbf{#1}}}
\scriptsize

\resizebox{\textwidth}{!}{%
\begin{tabular}{@{}l|*{10}{c}|c@{}}
\toprule
\textbf{Model}
& \textbf{MMSearch} & \textbf{MMSearch+} & \textbf{BC-VL} & \textbf{MM-BC} & \textbf{VisBrowse}
& \textbf{BC-V$^{3}$} & \textbf{VDR} & \textbf{SimpleVQA} & \textbf{LiveVQA} & \textbf{FVQA} & \textbf{Overall} \\

\midrule
\rowcolor{gray!10}
\multicolumn{12}{c}{\textbf{\textit{Direct Answer}}} \\
\midrule

GPT-5.6-Sol
& \textbf{68.3} & \underline{21.5} & \textbf{47.4} & \textbf{16.8} & \textbf{36.7}
& \textbf{22.7} & \textbf{19.2} & 69.3 & \textbf{73.0} & \textbf{63.7} & \textbf{43.9} \\

Gemini-3.1-Pro
& \underline{62.3} & \textbf{24.1} & \underline{41.6} & 11.8 & 24.3
& \underline{18.7} & \underline{16.0} & 64.7 & 64.3 & 57.7 & 38.5 \\

Kimi-K3
& 61.3 & 17.0 & 40.1 & 11.8 & \underline{30.8}
& 15.3 & 11.4 & \underline{70.0} & \underline{72.7} & \underline{61.0} & \underline{39.1} \\

Qwen3.8-Max
& 53.0 & 15.8 & 41.1 & \underline{14.0} & 28.4
& 11.7 & 8.8 & \textbf{72.7} & 68.7 & 59.0 & 37.3 \\

Qwen3.5-35B-A3B
& 26.7 & 4.5 & 25.3 & 8.5 & 17.8
& 6.3 & 2.6 & 55.3 & 56.3 & 34.3 & 23.8 \\

Qwen3.5-9B
& 22.3 & 1.9 & 24.8 & 6.8 & 17.2
& 5.0 & 2.2 & 52.7 & 54.7 & 28.3 & 21.6 \\

\midrule
\rowcolor{gray!10}
\multicolumn{12}{c}{\textbf{\textit{Agentic Workflow}}} \\
\midrule

GPT-5.6-Sol
& \textbf{88.7} & \textbf{65.9} & \textbf{66.4} & \textbf{59.5} & \textbf{70.4}
& \underline{55.7} & \textbf{44.6} & \textbf{87.0} & \textbf{94.0} & \textbf{88.7} & \textbf{72.1} \\

Gemini-3.1-Pro
& 85.0 & 61.1 & 61.7 & 55.3 & 63.3
& 51.3 & \underline{40.2} & 83.0 & 88.7 & 81.3 & 67.1 \\

Kimi-K3
& 82.3 & \underline{65.0} & \underline{64.7} & \underline{56.5} & \underline{68.0}
& 53.3 & 33.8 & 85.0 & 91.0 & 82.0 & 68.1 \\

Qwen3.8-Max
& \underline{85.7} & 63.3 & \textbf{66.4} & 54.5 & 66.9
& \textbf{56.3} & 39.0 & \underline{85.7} & \underline{93.0} & \underline{86.7} & \underline{69.7} \\

Qwen3.5-35B-A3B
& 71.0 & 28.0 & 51.4 & 29.0 & 42.0
& 30.7 & 19.4 & 71.7 & 83.7 & 65.0 & 49.2 \\

Qwen3.5-9B
& 66.3 & 24.8 & 43.6 & 22.8 & 33.7
& 22.3 & 17.0 & 66.7 & 74.7 & 61.3 & 43.3 \\

\midrule
\rowcolor{gray!10}
\multicolumn{12}{c}{\textbf{\textit{Multimodal Deep Search Agents}}} \\
\midrule

MMSearch-R1-7B
& 53.8 & -- & -- & -- & --
& 4.0 & -- & 57.4 & 48.4 & 58.4 & -- \\

DeepEyesV2-7B
& 63.7 & -- & -- & -- & --
& -- & -- & 59.4 & -- & 60.6 & -- \\

Vision-DeepResearch-8B
& 69.6 & 20.4 & 42.6 & -- & --
& 11.7 & 29.2 & -- & 76.7 & 64.7 & -- \\

MM-DeepResearch-8B
& 67.8 & -- & 37.9 & -- & --
& -- & -- & 65.9 & 65.0 & 69.2 & -- \\

POINTS-Seeker-8B
& 70.8 & 25.2 & 44.4 & -- & --
& -- & -- & 68.8 & 77.7 & 71.2 & -- \\

OpenSearch-VL-8B
& 64.5 & -- & 37.6 & -- & --
& -- & 20.8 & 71.6 & 59.6 & 71.5 & -- \\

Visual-Seeker-8B
& 72.2 & 27.3 & 47.6 & 16.1 & 34.7
& -- & -- & -- & -- & -- & -- \\

SearchEyes-9B
& 69.2 & -- & 42.3 & -- & --
& -- & 28.3 & 75.4 & 66.1 & 74.2 & -- \\

SimpleSearch-VL-8B
& 77.1 & 32.5 & 52.1 & -- & --
& -- & -- & 76.6 & 75.2 & 76.8 & -- \\

DeepVoyager-VL-8B
& 72.7 & 37.1 & 58.4 & 24.0 & 47.3
& 32.3 & \textbf{35.0} & 76.3 & 82.7 & \underline{82.7} & 54.9 \\

\rowcolor{blue!6}
\textbf{Neos-9B (Ours, SFT)}
& \underline{79.7} & \underline{44.7} & \underline{60.4} & \underline{37.8} & \underline{49.1}
& \underline{35.3} & 30.2 & \textbf{82.3} & \underline{87.7} & 79.7 & \underline{58.7} \\
\noalign{\vskip-0.7pt}
\rowcolor{blue!6}
\mmdgain{\quad $\Delta_{\mathrm{SFT}}$ vs. 9B workflow}
& \mmdgain{+13.4} & \mmdgain{+19.9} & \mmdgain{+16.8} & \mmdgain{+15.0} & \mmdgain{+15.4}
& \mmdgain{+13.0} & \mmdgain{+13.2} & \mmdgain{+15.6} & \mmdgain{+13.0} & \mmdgain{+18.4} & \mmdgain{+15.4} \\[-1.2pt]

\rowcolor{blue!6}
\textbf{Neos-9B (Ours, RL)}
& \textbf{81.3} & \textbf{47.6} & \textbf{63.2} & \textbf{39.5} & \textbf{53.3}
& \textbf{37.0} & \underline{31.4} & \underline{81.3} & \textbf{88.7} & \textbf{83.3} & \textbf{60.7} \\
\noalign{\vskip-0.7pt}
\rowcolor{blue!6}
\mmdgain{\quad $\Delta_{\mathrm{RL}}$ vs. 9B workflow}
& \mmdgain{+15.0} & \mmdgain{+22.8} & \mmdgain{+19.6} & \mmdgain{+16.7} & \mmdgain{+19.6}
& \mmdgain{+14.7} & \mmdgain{+14.4} & \mmdgain{+14.6} & \mmdgain{+14.0} & \mmdgain{+22.0} & \mmdgain{+17.4} \\[-1.2pt]

\midrule

WebWatcher-32B
& 55.3 & -- & 27.0 & -- & --
& 8.7 & -- & 59.0 & 58.7 & -- & -- \\

SenseNova-MARS-32B
& 74.3 & -- & -- & -- & --
& -- & -- & 74.1 & 60.8 & 72.6 & -- \\

Vision-DeepResearch-30B-A3B
& 69.6 & 28.5 & 53.7 & -- & --
& -- & \underline{37.8} & -- & 77.6 & 74.2 & -- \\

REDSearcher-MM-30B-A3B
& 72.9 & 26.6 & 57.2 & 23.5 & --
& -- & -- & -- & 79.3 & -- & -- \\

MM-DeepResearch-32B
& 69.0 & -- & 43.0 & -- & --
& -- & -- & 67.6 & 68.0 & 70.1 & -- \\

LMM-Searcher-30B-A3B
& 72.3 & 34.8 & -- & 30.1 & 48.3
& -- & -- & -- & -- & -- & -- \\

OpenSearch-VL-32B
& 72.3 & -- & 43.8 & -- & --
& -- & 33.8 & 76.2 & 70.5 & 74.7 & -- \\

SearchEyes-27B
& 82.4 & -- & 49.3 & -- & --
& -- & \textbf{39.4} & 80.9 & 77.3 & 79.1 & -- \\

SimpleSearch-VL-30B-A3B
& \underline{83.6} & 34.4 & 55.9 & -- & --
& -- & -- & 79.6 & 81.1 & 79.0 & -- \\

UniBrowse-35B-A3B
& 75.2 & 34.1 & 57.8 & 28.2 & --
& -- & -- & -- & -- & 76.7 & -- \\

DeepVoyager-VL-30B-A3B
& 74.0 & 40.6 & \underline{64.2} & 30.5 & \underline{53.8}
& 35.0 & \textbf{39.4} & 81.0 & 82.7 & \textbf{84.7} & 58.6 \\

\rowcolor{blue!6}
\textbf{Neos-35B (Ours, SFT)}
& 81.0 & \underline{48.6} & 61.4 & \underline{44.8} & 52.7
& \underline{45.3} & 31.0 & \underline{81.3} & \underline{89.3} & 79.7 & \underline{61.5} \\
\noalign{\vskip-0.7pt}
\rowcolor{blue!6}
\mmdgain{\quad $\Delta_{\mathrm{SFT}}$ vs. 35B workflow}
& \mmdgain{+10.0} & \mmdgain{+20.6} & \mmdgain{+10.0} & \mmdgain{+15.8} & \mmdgain{+10.7}
& \mmdgain{+14.6} & \mmdgain{+11.6} & \mmdgain{+9.6} & \mmdgain{+5.6} & \mmdgain{+14.7} & \mmdgain{+12.3} \\[-1.2pt]

\rowcolor{blue!6}
\textbf{Neos-35B (Ours, RL)}
& \textbf{83.7} & \textbf{52.7} & \textbf{66.4} & \textbf{48.3} & \textbf{57.4}
& \textbf{47.0} & 33.6 & \textbf{82.7} & \textbf{91.3} & \underline{83.7} & \textbf{64.7} \\
\noalign{\vskip-0.7pt}
\rowcolor{blue!6}
\mmdgain{\quad $\Delta_{\mathrm{RL}}$ vs. 35B workflow}
& \mmdgain{+12.7} & \mmdgain{+24.7} & \mmdgain{+15.0} & \mmdgain{+19.3} & \mmdgain{+15.4}
& \mmdgain{+16.3} & \mmdgain{+14.2} & \mmdgain{+11.0} & \mmdgain{+7.6} & \mmdgain{+18.7} & \mmdgain{+15.5} \\[-1.2pt]

\bottomrule
\end{tabular}%
}
\endgroup
\vspace{-4pt}
\end{table*}

\section{Experiments}
\label{sec:experiments}

\subsection{Experimental Setup}
\label{sec:experimentsetup}

\textbf{Benchmarks.}
We evaluate Neos on ten benchmarks spanning factual visual question answering, multi-step information search, and reasoning over online visual evidence.
The evaluation suite comprises MMSearch~\citep{jiang2024mmsearch}, SimpleVQA~\citep{cheng2025simplevqa}, LiveVQA~\citep{fu2025livevqa}, FVQA~\citep{wu2025mmsearch}, BrowseComp-VL~\citep{geng2026webwatcher}, MM-BrowseComp~\citep{li2025mm}, MMSearch-Plus~\citep{tao2026mmsearch}, VDR-Bench~\citep{zeng2026vision}, BrowseComp-V$^3$~\citep{zhang2026browsecomp}, and VisBrowse-Bench~\citep{zhang2026visbrowse}.

\textbf{Baselines.}
We evaluate three categories of baselines, covering proprietary and open-source MLLMs as well as specialized multimodal search agents (Table~\ref{tab:main_results}).
\textit{Direct Answer} measures model capability using only the input question and images, without external tools.
\textit{Agentic Workflow} allows the same models to gather textual evidence and inspect images through our tool framework, assessing the benefit of iterative search without additional model training.
\textit{Multimodal deep search Agents} compares against existing systems specifically trained for multimodal search and browsing, providing a comparison with learned search policies.

\textbf{Implementation Details.}
We perform cold-start SFT of Qwen3.5-9B and Qwen3.5-35B-A3B~\citep{qwen35} using MS-Swift~\citep{zhao2025swift} on 8 NVIDIA H100 GPUs for two epochs.
Agentic RL is implemented in verl~\citep{sheng2025hybridflow} with Megatron~\citep{shoeybi2019megatron} as the training backend.
The co-evolution stage uses 40 H100 GPUs, with 32 allocated to agentic RL and 8 to MEI serving and fine-tuning.
Policy RL is performed for one epoch with 8 rollouts per question, using a batch size of 64 for Neos-9B and 32 for Neos-35B.
The vision encoder and multimodal merger remain frozen throughout training, and evaluation allows up to 50 interaction turns per question.
Detailed experimental settings are provided in Appendix~\ref{appendix:experimental_details}.

\begin{table*}[t]
\centering
\begingroup

\setlength{\parskip}{0pt}
\captionsetup{
  position=top,
  skip=0pt,
  belowskip=0pt
}

\def\NeosLeftWidth{0.52\textwidth}
\def\NeosRightWidth{0.46\textwidth}

\begin{minipage}[t]{\NeosLeftWidth}
\vspace{0pt}
\caption{
\textbf{Training-data ablation.}
The VQA mixture draws from FVQA, LiveVQA, InfoSeek, and VisualProbe,
with the same expert model for trajectory generation and rejection sampling.
}
\label{tab:ablation_data}
\end{minipage}\hfill
\begin{minipage}[t]{\NeosRightWidth}
\vspace{0pt}
\caption{
\textbf{Co-evolution ablation.}
Initial and adapted refer to MEI before and after co-evolution.
The RL settings use the same SFT policy initialization.
}
\label{tab:ablation_coevolution}
\end{minipage}\par

\nointerlineskip
\vspace{4pt}

\fontsize{8}{9.2}\selectfont
\setlength{\tabcolsep}{1.5pt}
\renewcommand{\arraystretch}{1.0}
\setlength{\aboverulesep}{1.2pt}
\setlength{\belowrulesep}{1.6pt}

\def\NeosDataBody#1{%
  \begin{tabularhtx}{#1}{\linewidth}
  {@{}>{\raggedright\arraybackslash}Xccc@{}}
  \toprule
  \textbf{SFT data}\vphantom{\textbf{BC-V$^{3}$}}
  & \textbf{BC-VL} & \textbf{MMSearch+} & \textbf{FVQA} \\
  \midrule
  \interrowfill

  Base Qwen3.5-9B
  & 43.6 & 24.8 & 61.3 \\
  \interrowfill

  \midrule
  \interrowfill
  VQA mixture (0.7K)
  & 53.4 & 32.9 & 72.7 \\
  \interrowfill
  \interrowfill

  REDSearcher-MM (1.1K)
  & 56.9 & 38.9 & 77.3 \\
  \interrowfill
  \interrowfill

  Our synthetic data (1K)
  & 58.4 & 43.2 & 78.3 \\
  \interrowfill

  \midrule
  \textbf{Full mixture (2.9K)}
  & \textbf{60.4} & \textbf{44.7} & \textbf{79.7} \\
  \bottomrule
  \end{tabularhtx}%
}

\def\NeosCoevolutionBody#1{%
  \begin{tabularhtx}{#1}{\linewidth}
  {@{}l>{\raggedright\arraybackslash}Xccc@{}}
  \toprule
  \textbf{Policy} & \textbf{MEI}
  & \textbf{MMSearch+} & \textbf{BC-V$^{3}$} & \textbf{VDR} \\
  \midrule
  \interrowfill

  SFT & Initial
  & 42.8 & 33.7 & 28.0 \\
  \interrowfill
  \interrowfill

  SFT & Adapted
  & 44.7 & 35.3 & 30.2 \\
  \interrowfill

  \midrule
  \interrowfill

  RL & Frozen initial
  & 45.8 & 35.7 & 28.2 \\
  \interrowfill
  \interrowfill

  RL & Frozen adapted
  & 46.9 & 36.3 & 30.4 \\
  \interrowfill

  \midrule
  \multicolumn{2}{@{}l}{\textbf{Alternating}}
  & \textbf{47.6} & \textbf{37.0} & \textbf{31.4} \\
  \bottomrule
  \end{tabularhtx}%
}

\setbox0=\hbox{%
  \begin{minipage}{\NeosLeftWidth}
    \NeosDataBody{spread=0pt}
  \end{minipage}%
}
\edef\NeosNaturalHeight{\the\dimexpr\ht0+\dp0\relax}

\setbox2=\hbox{%
  \begin{minipage}{\NeosRightWidth}
    \NeosCoevolutionBody{spread=0pt}
  \end{minipage}%
}

\dimen0=\dimexpr\ht2+\dp2\relax
\ifdim\dimen0<\NeosNaturalHeight\relax
  \dimen0=\NeosNaturalHeight\relax
\fi
\advance\dimen0 by 1pt
\edef\NeosTableHeight{\the\dimen0}

\noindent
\begin{minipage}[t]{\NeosLeftWidth}
\vspace{0pt}
\centering
\NeosDataBody{\NeosTableHeight}
\end{minipage}\hfill
\begin{minipage}[t]{\NeosRightWidth}
\vspace{0pt}
\centering
\NeosCoevolutionBody{\NeosTableHeight}
\end{minipage}\par

\endgroup

\vspace{-8pt}
\end{table*}

\subsection{Main Results}
\label{sec:main_results}



As shown in Table~\ref{tab:main_results}, Neos achieves leading performance among specialized multimodal deep-search agents at both model scales.
The advantage is particularly pronounced for Neos-35B on MMSearch-Plus, MM-BrowseComp, and BrowseComp-V$^{3}$, where it achieves 52.7, 48.3, and 47.0, surpassing the strongest prior specialized-agent results by 12.1, 17.8, and 12.0 points.
It also reaches 57.4 on VisBrowse-Bench and 91.3 on LiveVQA, demonstrating strong capabilities across both multimodal browsing and visual question answering.

The training progression further highlights the value of specialized training beyond merely providing tool access.
With our workflow, the Qwen3.5-9B and Qwen3.5-35B-A3B backbones improve from 21.6 and 23.8 to 43.3 and 49.2.
The proposed training pipeline effectively strengthens the models' interactive search, tool-use, and reasoning capabilities, yielding final scores of 60.7 and 64.7.
These correspond to gains of 17.4 and 15.5 points over the corresponding base workflows.
Neos-9B already outperforms several 30B-scale search agents, while the stronger results of Neos-35B show that effective agent training and larger model capacity are complementary.
Overall, Neos approaches frontier-model performance, demonstrating that specialized multimodal search agents can be made highly competitive with frontier workflows.

\subsection{Ablation Studies}
\label{sec:ablation}

Unless otherwise specified, ablations follow the settings of the corresponding stage in our main experiments.
Except for the MEI-related ablations, all experiments use our final co-evolved MEI. Additional ablations and analyses are provided in Appendix~\ref{appendix:ablations}.

\textbf{Training data.}
Table~\ref{tab:ablation_data} compares SFT with different training sources.
All trajectories, including those based on REDSearcher-MM and the VQA mixture, are generated with the same expert model.
Our 1K synthetic examples yield 58.4, 43.2, and 78.3 on BrowseComp-VL, MMSearch-Plus, and FVQA, exceeding the REDSearcher-MM variant by 1.5, 4.3, and 1.0 points.
Combining all sources further improves performance to 60.4, 44.7, and 79.7, suggesting complementary benefits from general visual question answering and structured multimodal search supervision.

\textbf{Agent--environment co-evolution.}
Table~\ref{tab:ablation_coevolution} separates the benefit of a stronger MEI from that of adapting it alongside the policy.
Beyond evaluating the SFT policy with different MEI checkpoints, we compare three RL settings from the same SFT policy initialization with matched policy-update budgets: freezing the initial MEI, freezing the co-evolved MEI, and alternating policy RL with MEI SFT.
Using the co-evolved MEI as a fixed interface improves over freezing the initial MEI by 1.1, 0.6, and 2.2 points on MMSearch-Plus, BrowseComp-V$^{3}$, and VDR-Bench.
Alternating updates further improve performance by 0.7, 0.7, and 1.0 points over this stronger fixed-interface control.
These results show that a better MEI accounts for part of the improvement, while continued adaptation during policy optimization provides additional gains across all three benchmarks.

\begin{figure}[h]
    \centering
    \includegraphics[width=1\linewidth]{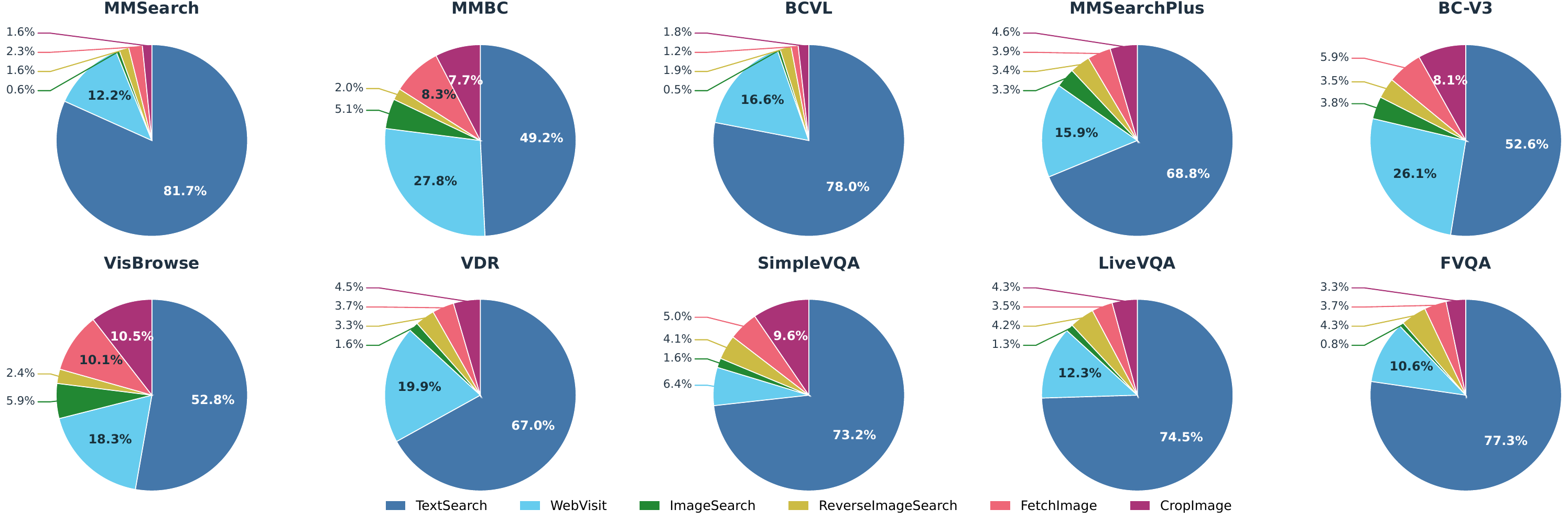}
    \caption{\textbf{Tool call distribution of ten benchmarks.}}
    \label{fig:tool_usage}
\end{figure}

\textbf{MEI backbone.}
We fix the final Neos-9B policy and vary the model serving as MEI in Table~\ref{tab:ablation_mei}.
Relative to the Qwen3.5-9B MEI baseline, our co-evolved MEI improves MM-BrowseComp, BrowseComp-VL, and VisBrowse-Bench by 4.5, 5.1, and 6.6 points, respectively.
It outperforms Qwen-3.7-Flash, GPT-5.6-Luna, and GPT-5-nano on all three benchmarks, while remaining within 1.8 points of Gemini-3.7-Flash on each.
These results show that a task-adapted 9B evidence interface can support search performance competitive with strong off-the-shelf models.

\textbf{Reference-guided credit assignment.}
Table~\ref{tab:ablation_credit} progressively introduces the credit components.
Outcome-only GRPO falls below the cold-start SFT baseline on all three benchmarks, while productive-only credit provides mixed gains.
Adding first-error penalties improves over productive-only credit by 0.7, 1.5, and 3.0 points on BrowseComp-VL, MM-BrowseComp, and FVQA, respectively.
Downstream attribution and corrective credit then yield further improvements across all three benchmarks.
The full method reaches 63.2, 39.5, and 83.3, exceeding outcome-only GRPO by 3.6, 4.2, and 5.0 points.
This progression highlights the value of distinguishing critical errors, downstream consequences, and useful actions rather than rewarding productive steps alone.

\begin{table*}[t]
\centering

\setlength{\parskip}{0pt}
\captionsetup{
  position=top,
  skip=0pt,
  belowskip=0pt
}

\begin{minipage}[t]{0.48\textwidth}
\vspace{0pt}
\caption{
\textbf{MEI backbone ablation.}
}
\label{tab:ablation_mei}
\end{minipage}\hfill
\begin{minipage}[t]{0.48\textwidth}
\vspace{0pt}
\caption{
\textbf{Credit-assignment ablation.}
}
\label{tab:ablation_credit}
\end{minipage}

\par\vspace{3pt}

\begingroup
\fontsize{8}{9.2}\selectfont
\setlength{\tabcolsep}{1.2pt}
\renewcommand{\arraystretch}{1.0}

\setlength{\aboverulesep}{1.4pt}
\setlength{\belowrulesep}{1.8pt}

\newcommand{\ablheight}{3.3cm}
\newcommand{\ablplus}{\hspace{0.4em}$+$\,}

\begin{minipage}[t]{0.48\textwidth}
\vspace{0pt}
\centering
\begin{tabularhtx}{\ablheight}{\linewidth}
{@{}>{\raggedright\arraybackslash}Xccc@{}}

\toprule
\textbf{MEI model}
& \textbf{MM-BC}
& \textbf{BC-VL}
& \textbf{VisBrowse} \\
\midrule
\interrowfill

Gemini-3.7-Flash
& 41.3 & 64.2 & 54.4 \\
\interrowfill

Qwen-3.7-Flash
& 39.3 & 62.4 & 51.5 \\
\interrowfill

GPT-5.6-Luna
& 38.7 & 61.2 & 52.7 \\
\interrowfill

GPT-5-nano
& 36.5 & 57.9 & 47.9 \\
\interrowfill

Qwen3.5-9B
& 35.0 & 58.1 & 46.7 \\
\interrowfill

\midrule
\textbf{Co-evolved MEI (9B)}
& 39.5 & 63.2 & 53.3 \\
\bottomrule

\end{tabularhtx}
\end{minipage}\hfill
\begin{minipage}[t]{0.48\textwidth}
\vspace{0pt}
\centering
\begin{tabularhtx}{\ablheight}{\linewidth}
{@{}>{\raggedright\arraybackslash}Xccc@{}}

\toprule
\textbf{Training variant}
& \textbf{BC-VL}
& \textbf{MM-BC}
& \textbf{FVQA} \\
\midrule
\interrowfill

Cold-start SFT
& 60.4 & 37.8 & 79.7 \\
\interrowfill

Outcome-only GRPO
& 59.6 & 35.3 & 78.3 \\
\interrowfill

\midrule
\interrowfill

Productive credit only
& 60.4 & 36.3 & 77.7 \\
\interrowfill

\ablplus First-error penalty
& 61.2 & 37.8 & 80.7 \\
\interrowfill

\ablplus Downstream attribution
& 61.9 & 38.3 & 81.3 \\
\interrowfill

\ablplus Corrective credit
& 62.7 & 39.0 & 82.0 \\
\interrowfill

\midrule
\textbf{Full credit assignment}
& \textbf{63.2} & \textbf{39.5} & \textbf{83.3} \\
\bottomrule

\end{tabularhtx}
\end{minipage}

\endgroup

\vspace{-4pt}
\end{table*}

\begin{figure}[h]
    \centering
    \includegraphics[width=1\linewidth]{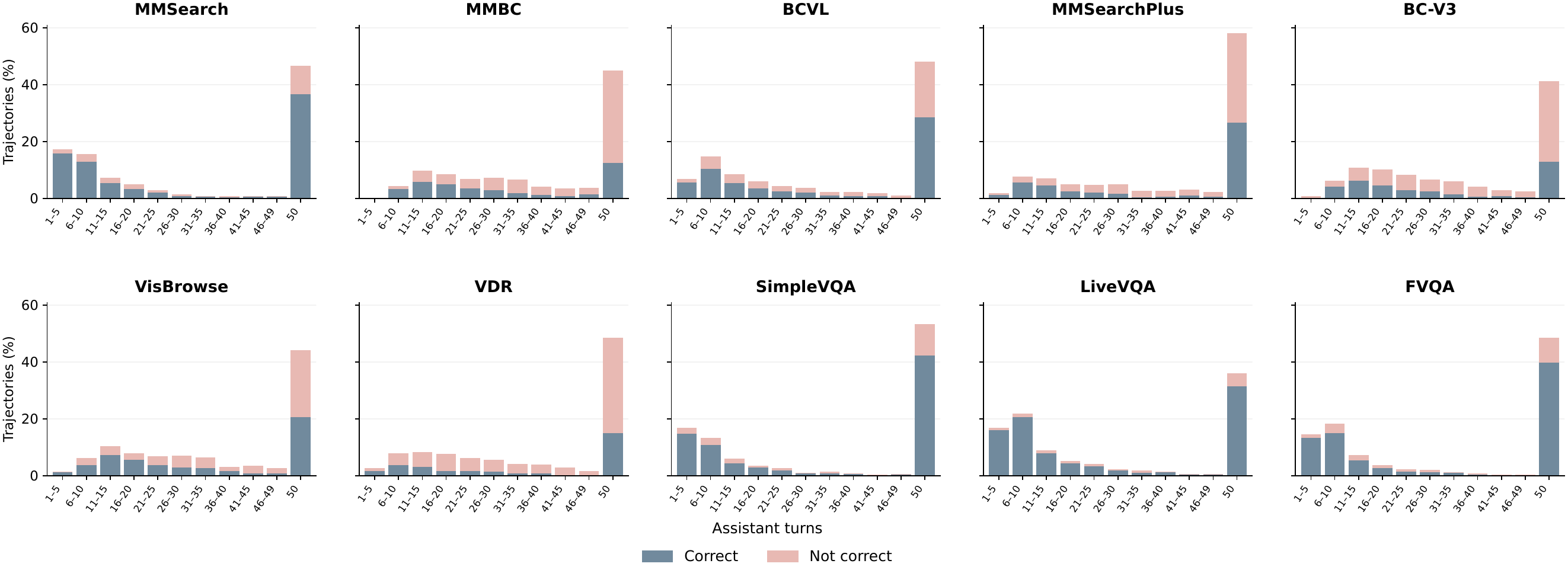}
    \caption{\textbf{Neos-9B's turn distribution across ten benchmarks.}}
    \label{fig:turn_distribution}
    \vspace{-10pt}
\end{figure}

\textbf{Tool-use analysis.}
Figure~\ref{fig:tool_usage} shows that \texttt{TextSearch} accounts for 81.7 and 78.0 of calls on MMSearch and BrowseComp-VL.
This prevalence may partly reflect the relatively low overhead of text-based exploration, together with online textual shortcuts that allow some benchmark questions to bypass visual search~\citep{zeng2026vision}.
In contrast, its share falls to 49.2, 52.6, and 52.8 on MM-BrowseComp, BrowseComp-V$^{3}$, and VisBrowse-Bench, where visual tools contribute approximately 23, 21, and 29.
These patterns suggest that more demanding visual investigations shift the tool mix toward image acquisition and inspection, beyond comparatively inexpensive textual exploration.

\begin{figure}[h]
    \centering
    \includegraphics[width=1\linewidth]{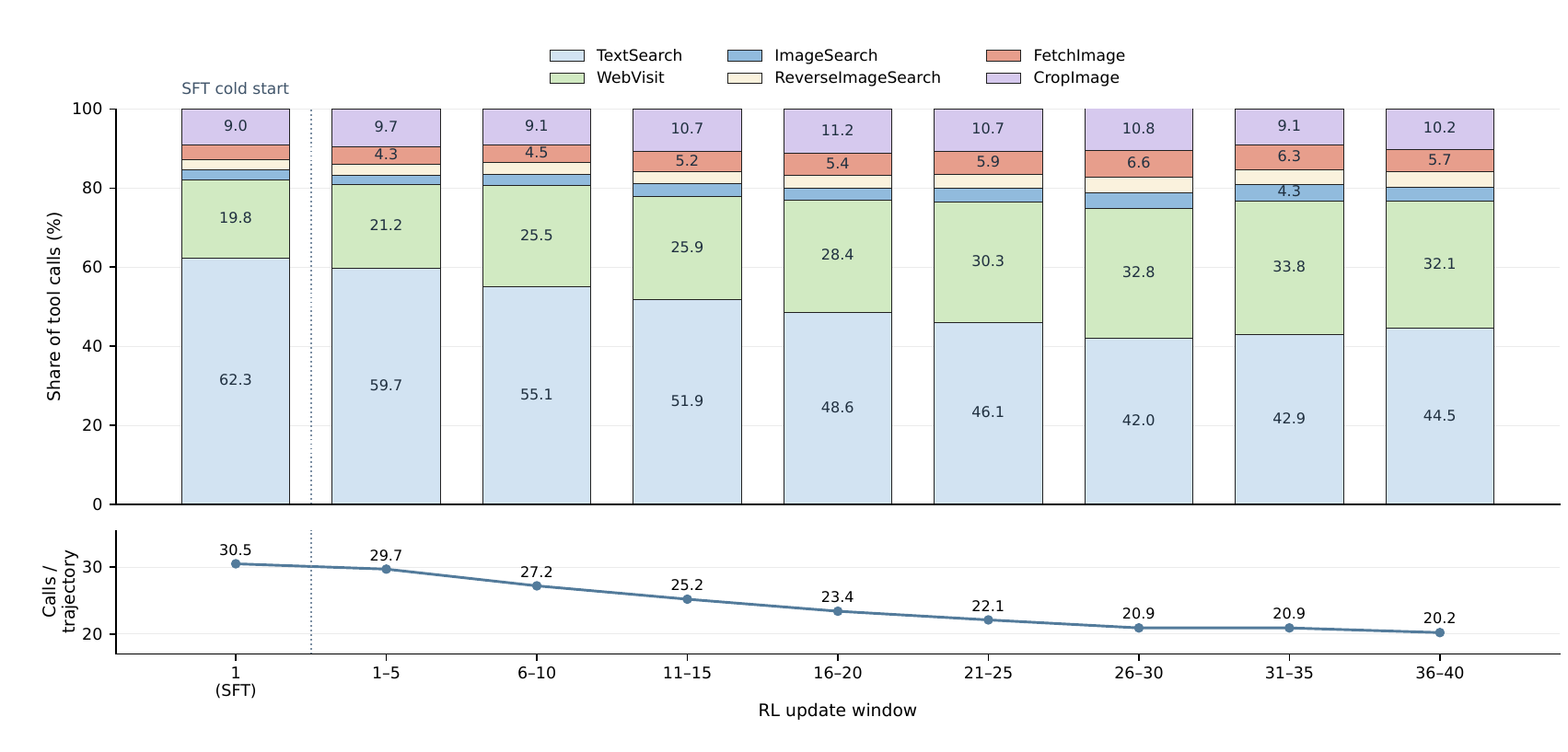}
    \caption{\textbf{Tool usage distribution during Neos-9B's co-evolution training.}}
    \label{fig:rl_tool_usage}
    \vspace{-18pt}
\end{figure}

\paragraph{Search horizons across benchmarks.}
Figure~\ref{fig:turn_distribution} shows Neos-9B's interaction lengths, separated by answer correctness.
Before the 50-turn limit, MMSearch and the VQA benchmarks concentrate in the first ten turns, whereas others span broader horizons, consistent with more sustained multimodal investigation.
All benchmarks nevertheless accumulate at the limit; these trajectories are predominantly correct on LiveVQA and FVQA but predominantly incorrect on MM-BrowseComp, BrowseComp-V$^{3}$, and VDR-Bench.
Thus, search lengths are heterogeneous, and reaching the budget alone indicates neither success nor failure.

\textbf{Tool-use dynamics during RL.}
Figure~\ref{fig:rl_tool_usage} tracks the composition and frequency of tool calls across RL update windows.
From the first to the final window, the share of \texttt{TextSearch} decreases from 59.7 to 44.5, while \texttt{WebVisit} increases from 21.2 to 32.1.
The four visual tools---\texttt{ImageSearch}, \texttt{ReverseImageSearch}, \texttt{FetchImage}, and \texttt{CropImage}---collectively increase their share from approximately 19.1 to 23.4.
Meanwhile, the average number of calls per trajectory falls from 29.7 to 20.2, a reduction of 32.0\%.
Thus, the changing tool mix accompanies fewer overall calls rather than an expansion of interaction volume.
These trends are consistent with a shift from text-search-heavy exploration toward more selective information gathering, with greater relative emphasis on examining source pages and visual evidence.
The resulting trajectories are more compact in tool-call count while retaining a diverse set of evidence-processing operations.

\textbf{Tool ablation.}
Table~\ref{tab:ablation_tools} evaluates individual tool contributions using Qwen3.5-35B-A3B without additional search-specific policy training; removing any tool hurts all three benchmarks, with task-dependent effects.
\texttt{ReverseImageSearch} is most critical on SimpleVQA (71.7 to 58.0), where removing \texttt{ImageSearch} also costs 4.7 points.
By contrast, \texttt{FetchImage} causes the largest drops on BrowseComp-V$^{3}$ and MM-BrowseComp (3.7 points each), while \texttt{CropImage} and \texttt{WebVisit} also contribute consistently.
Together, these results support combining evidence discovery, page reading, and visual inspection in the Neos workflow.

\begin{table*}[t]
\centering
\begingroup

\def\NeosToolWidth{0.46\textwidth}
\def\NeosWorkflowWidth{0.52\textwidth}
\def\NeosAblHeight{2.9cm}
\setlength{\parskip}{0pt}

\begin{minipage}[t]{\NeosToolWidth}
\vspace{0pt}
\caption{
\textbf{Tool ablation using Qwen3.5-35B-A3B base model.}
}
\label{tab:ablation_tools}
\end{minipage}\hfill
\begin{minipage}[t]{\NeosWorkflowWidth}
\vspace{0pt}
\caption{
\textbf{Workflow comparison with our reproduction of REDSearcher-MM.}
}
\label{tab:ablation_workflow}
\end{minipage}

\par\vspace{2pt}

\fontsize{8}{9.2}\selectfont
\setlength{\tabcolsep}{1.2pt}
\renewcommand{\arraystretch}{1.05}
\setlength{\aboverulesep}{1.2pt}
\setlength{\belowrulesep}{1.6pt}

\noindent
\begin{minipage}[t]{\NeosToolWidth}
\vspace{0pt}
\centering

\begin{tabularht*}{\NeosAblHeight}{\linewidth}
{@{\extracolsep{\fill}}lccc@{}}
\toprule
\textbf{Tool configuration}
& \textbf{SimpleVQA}
& \textbf{BC-V$^{3}$}
& \textbf{MM-BC} \\
\midrule

\textbf{Full Neos workflow}
& \textbf{71.7}
& \textbf{30.7}
& \textbf{29.0} \\

\midrule
\interrowfill

w/o FetchImage
& 68.7 & 27.0 & 25.3 \\
\interrowfill

w/o CropImage
& 69.7 & 29.0 & 28.3 \\
\interrowfill

w/o WebVisit
& 70.3 & 29.0 & 27.0 \\
\interrowfill

w/o ImageSearch
& 67.0 & 28.7 & 27.7 \\
\interrowfill

w/o ReverseImageSearch
& 58.0 & 27.3 & 28.3 \\

\bottomrule
\end{tabularht*}
\end{minipage}\hfill
\begin{minipage}[t]{\NeosWorkflowWidth}
\vspace{0pt}
\centering

\begin{tabularht*}{\NeosAblHeight}{\linewidth}
{@{\extracolsep{\fill}}llccc@{}}
\toprule
\textbf{Policy}\vphantom{\textbf{BC-V$^{3}$}}
& \textbf{Workflow}
& \textbf{BC-VL}
& \textbf{MM-BC}
& \textbf{MMSearch+} \\
\midrule

\multirow{2}{*}{GPT-5.6}
& REDSearcher-MM
& 64.2 & 35.5 & 53.5 \\

& Neos
& \textbf{66.4} & \textbf{59.5} & \textbf{65.9} \\

\interrowfill
\midrule
\interrowfill

\multirow{2}{*}{\shortstack[l]{Qwen3.5-\\35B-A3B}}
& REDSearcher-MM
& 45.4 & 14.5 & 21.6 \\

& Neos
& \textbf{51.4} & \textbf{29.0} & \textbf{28.0} \\

\interrowfill
\midrule
\interrowfill

\multirow{2}{*}{\shortstack[l]{Qwen3.5-\\9B}}
& REDSearcher-MM
& 38.6 & 9.8 & 19.3 \\

& Neos
& \textbf{43.6} & \textbf{22.8} & \textbf{24.8} \\

\bottomrule
\end{tabularht*}
\end{minipage}

\par
\endgroup

\vspace{-6pt}
\end{table*}

\textbf{Workflow comparison.}
Table~\ref{tab:ablation_workflow} compares Neos with our reproduction of the REDSearcher-MM workflow~\citep{chu2026redsearcher}, holding the policy model fixed within each comparison.
Neos improves performance across all nine model--benchmark pairs, with average gains of 12.9, 9.0, and 7.8 points across the three benchmarks for GPT-5.6, Qwen3.5-35B-A3B, and Qwen3.5-9B.
The largest improvements occur on MM-BrowseComp, reaching 24.0, 14.5, and 13.0 points.
Notably, Qwen3.5-9B with Neos surpasses Qwen3.5-35B-A3B with REDSearcher-MM on both MM-BrowseComp and MMSearch-Plus.
These results indicate that the benefits of Neos extend across policy backbones and that workflow design can partially compensate for differences in model scale.

\begin{figure}[t]
\centering
\includegraphics[width=0.78\linewidth]{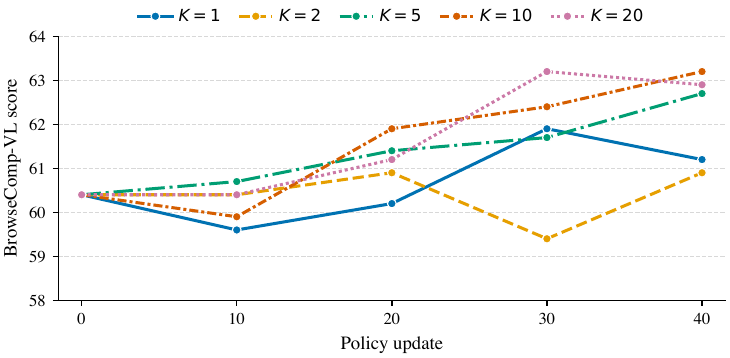}
\caption{
\textbf{Effect of the MEI update interval on BrowseComp-VL with the 9B model.}
$K$ is the number of policy-training batches between MEI updates.
}
\label{fig:ablation_mei_interval}
\vspace{-8pt}
\end{figure}

\textbf{MEI update interval.}
Figure~\ref{fig:ablation_mei_interval} shows that frequent MEI updates with small $K$ are unstable, producing oscillations and occasional regressions across policy-training checkpoints.
As $K$ increases, performance becomes progressively more stable, but the gains are neither monotonic nor linear in the update interval.
This suggests that batching more policy updates before refreshing MEI primarily stabilizes policy--interface co-evolution rather than yielding proportional performance improvements.

\section{Conclusions}
We presented \textbf{Neos}, a multimodal deep search framework that sustains an \emph{evidence--decision loop} by learning both evidence acquisition and evidence-conditioned decision making.
Its evidence-dependent pipeline constructs tasks with intermediate visual dependencies while screening textual shortcuts.
A lightweight Multimodal Evidence Interface (MEI) co-adapts with the policy through agent--environment co-evolution, while reference-guided credit assignment localizes critical errors and preserves credit for useful decisions in failed searches.
Alternating policy RL with MEI SFT turns rollout outcomes into supervision for both policy and MEI.
Across ten benchmarks, Neos-9B and Neos-35B achieve leading performance among specialized multimodal deep search agents and substantially narrow the gap to frontier proprietary models.

\clearpage

\bibliographystyle{plainnat}
\bibliography{ref}

\clearpage

\appendix

\section{Details of Our Data Pipeline and Training Data}
\label{appendix:data}

\subsection{Details of Our Data Pipeline}
We construct questions designed to require newly acquired visual evidence during multimodal deep search.
Their intended solutions interleave relation search with visual grounding and, when needed, visual answer extraction.
Intermediate images resolve entities needed for subsequent search, while images of the resolved target provide visual attributes requested by the question.
These roles can coexist within a question and recur along its solution path.
Our pipeline first selects seed entities and constructs relational reasoning graphs, then derives and verifies programs with explicit visual dependencies, grounds them in online image evidence, and realizes natural-language questions before filtering candidates and curating successful search trajectories.

\paragraph{Selecting seeds for compositional visual search.}
Seed selection considers both the relational structure around an entity and the opportunities it offers for visual grounding.
We draw from Wikipedia-linked Wikidata entities with substantive encyclopedic coverage, excluding disambiguation pages and non-entity entries.
Within semantic domains, we prioritize diverse incoming and outgoing relations and the presence of image-associated neighbors, while downweighting highly connected hubs and neighborhoods dominated by generic attributes or taxonomic links.
This favors seeds whose neighborhoods offer several potential visual checkpoints along a reasoning structure, including entities other than the seed itself.
These signals guide construction; successful visual grounding and resistance to textual shortcuts are assessed downstream.

\paragraph{Expanding seeds into relational DAGs.}
We treat each seed as a candidate terminal entity and expand its relational prerequisites backward.
For each expanded entity, we select a small \emph{distinguishing set} of neighboring entities and relations whose conjunction uniquely identifies it in the KG snapshot.
The neighbors become prerequisites for resolving that entity, and recursively expanding them produces a directed acyclic graph (DAG) of identification dependencies.
Edges are oriented from prerequisites to the entity they identify, with the direction of each underlying Wikidata fact preserved.
Expansion explores complementary relations across branches and recursively extends their dependencies, subject to depth and branching constraints.
We favor concrete semantic relations, such as creation, membership, and location, and exclude encyclopedic maintenance and bibliographic-source links.
The resulting DAG provides a structured search space for extracting reasoning structures, with entity identities and relation constraints established before question realization.

\paragraph{Constructing visual dependencies.}
From each DAG, we enumerate paths leading to the seed and subgraphs with two branches that converge on a shared entity, which either is the terminal entity or leads to it through further relations.
Convergent branches have disjoint entity sets before the merge and constrain the shared entity through distinct relations.
We select an upstream entity as the visual entry anchor and identify feasible visual checkpoints within the program.
At an intermediate checkpoint, the visually grounded entity supplies the input to a subsequent relation; at an answer checkpoint, the retrieved image supplies the requested property of the resolved target.
Programs contain one or multiple intermediate checkpoints, optionally followed by visual answer extraction.
We apply structural filters before acquiring images and compile each program into a conjunctive query over the complete local KG snapshot (Equation~\ref{eq:data_identifiability}), where $C$ specifies relation constraints expressed in their original KG direction, $B$ specifies entry and planned visual-entity bindings, and $z_{\mathrm{out}}$ denotes the terminal entity.
For a seed entity $u^\star$, we retain programs with $\mathcal U_{\mathcal K}(C,B)=\{u^\star\}$; multiple endpoints or unresolved query evaluations are rejected.
This verifies endpoint identity conditional on the planned bindings.
The target is either $u^\star$ or a visual property subsequently verified from images of that target.

\paragraph{Grounding dependencies in online evidence.}
For each verified program, a planner discovers images and source pages for the required checkpoints without introducing new graph relations.
A blind observer describes full images using sanitized scene cues, with target names, search titles, graph relations, and expected answers withheld.
A separate observer examines actual crops without receiving the first observer's conclusions.
For entity grounding, a binding step matches the visual evidence to candidate entities including distractors, without marking the intended target.
We accept identities supported by distinctive pixels or by compatible visual observations corroborated through reverse search or source pages; incorrect and unresolved matches are rejected.
Each intermediate binding is explicitly connected to its subsequent relation.
For visual answer extraction, both full-image and crop observations must support the requested attribute.
When a program spans converging branches, we additionally check distinct images, different source domains, and consistency with the verified graph.
We verbalize the relations, compose a base question, and rewrite it to remove revealing names and mechanical structural cues while preserving its facts and answer.
A semantic-coverage check verifies that each required relation remains expressed.
The solver receives only the question and entry image; images needed at later checkpoints, including final attribute reading, must be acquired during search.

\paragraph{Filtering and curating search trajectories.}
We sequentially test candidates using text-only answering, text-only search with \texttt{TextSearch}, direct visual question answering with the entry image, and short multimodal search.
Both search settings allow at most four tool-use turns followed by a mandatory final-answer turn.
Any correct answer rejects the candidate; unresolved infrastructure or judging failures are excluded.
These filters screen for shortcuts and insufficient search depth relative to the chosen models and budgets.
For surviving questions, a stronger teacher performs long-horizon search, and an independent answer judge selects successful trajectories.
We retain traces with complete observations, valid action order, and compliant sequence lengths for policy SFT and training-side success references.
Local tool tasks additionally provide MEI training inputs, with targets separately regenerated and checked against the supplied material.

\subsection{Training Data}
\label{appendix:training_data}

We construct the policy training sets from our synthetic search tasks, REDSearcher-MM data, and a mixture of visual question answering datasets.
Qwen3.8-Max serves as the expert model for trajectory generation and rejection sampling within the Neos workflow.
Table~\ref{tab:training_data_composition} summarizes the resulting SFT and RL datasets.

\paragraph{Data sources and expert trajectories.}
Our data pipeline produces approximately 2K synthetic search instances with verified successful trajectories.
To incorporate external search questions, we rank the REDSearcher-MM SFT data~\citep{chu2026redsearcher} by decreasing trajectory length, measured in interaction turns, and retain the top 60\%.
We additionally sample approximately 3.7K candidate questions from FVQA~\citep{wu2025mmsearch}, LiveVQA~\citep{fu2025livevqa}, InfoSeek~\citep{chen2023can}, and VisualProbe~\citep{lai2026mini}.
For FVQA and LiveVQA, all sampled instances are selected from their corresponding official training splits, and after deduplication against the benchmarks used in our evaluation, the overlap rate with the evaluation benchmarks is 0\%.
For these external questions, we generate new trajectories using Qwen3.8-Max in Neos and retain successful trajectories through rejection sampling, rather than directly adopting demonstrations produced under another workflow.

\paragraph{SFT composition.}
We allocate 989 examples from our synthetic data to policy SFT and reserve the remaining 973 instances for RL.
From the expert-generated external trajectories, we select 1,177 based on REDSearcher-MM questions and 732 based on the VQA mixture.
The resulting SFT dataset contains 2,898 trajectories, combining demonstrations of interleaved multimodal search with broader visual question answering.

\paragraph{RL question selection.}
After cold-start SFT, we generate three rollouts for each question in the available data pool using the initialized policy, excluding the 973 synthetic instances reserved for RL.
We retain questions answered correctly in zero or one of the three rollouts.
The selected pool contains 1,599 questions: 599 from our synthetic data, 500 from REDSearcher-MM, and 500 from the VQA mixture.
Combining these questions with the 973 reserved synthetic instances yields an RL dataset of 2,572 questions.
Table~\ref{tab:training_data_composition} summarizes the composition of both training sets.

\begin{table}[h]
\centering
\caption{
\textbf{Composition of the policy training datasets.}
SFT counts refer to expert-generated trajectories, whereas RL counts refer to questions.
Difficulty filtering retains questions answered correctly in zero or one of three cold-start policy rollouts.
}
\label{tab:training_data_composition}

\begingroup
\small
\setlength{\tabcolsep}{5pt}
\renewcommand{\arraystretch}{1.12}

\begin{tabularx}{\linewidth}{@{}
  >{\raggedright\arraybackslash}X
  >{\raggedright\arraybackslash}X
  c@{}}
\toprule
\textbf{Data source} & \textbf{Selection} & \textbf{Count} \\
\midrule

\textbf{SFT total}
& & \textbf{2,898} \\
\addlinespace[2pt]

\quad Our synthetic data
& Expert trajectories
& 989 \\

\quad REDSearcher-MM
& Expert trajectories
& 1,177 \\

\quad VQA mixture
& Expert trajectories
& 732 \\

\midrule

\textbf{RL total}
& & \textbf{2,572} \\
\addlinespace[2pt]

\quad Reserved synthetic data
& Reserved before difficulty filtering
& 973 \\

\quad Our synthetic data
& Difficulty-filtered
& 599 \\

\quad REDSearcher-MM
& Difficulty-filtered
& 500 \\

\quad VQA mixture
& Difficulty-filtered
& 500 \\

\bottomrule
\end{tabularx}

\endgroup
\end{table}

\begin{table*}[h]
  \centering
  \caption{\textbf{Tools exposed to Neos.} MEI is embedded in tools. Search and page-reading return text and image references; pixels enter the policy context through \texttt{FetchImage} or \texttt{CropImage}.}
  \label{tab:Neos-tools}
  \begingroup
  \small
  \setlength{\tabcolsep}{5pt}
  \renewcommand{\arraystretch}{1.18}
  \begin{tabularx}{\textwidth}{@{}l >{\raggedright\arraybackslash}X >{\raggedright\arraybackslash}X >{\raggedright\arraybackslash}X@{}}
    \toprule
    Tool & Role & Input & Output \\
    \midrule
    \texttt{TextSearch} & Discover candidate webpages & Text query or query batch & Page titles, URLs, and snippets \\
    \texttt{WebVisit} & Extract page evidence with MEI & Page URL(s) and local goal & Goal-relevant evidence and page-image URLs \\
    \texttt{ImageSearch} & Rank image candidates with MEI & Text query(s) and visual goal & Ranked image/source URLs and visual analysis \\
    \texttt{ReverseImageSearch} & Match visual evidence with MEI & Image IDs/URLs and matching goal & Match analysis and candidate/source URLs \\
    \texttt{FetchImage} & Load selected visual evidence & Image URLs; optional captions & Image pixels and reusable image IDs \\
    \texttt{CropImage} & Inspect fine-grained regions & Image IDs and normalized bounding boxes & Crop pixels and reusable image IDs \\
    \bottomrule
  \end{tabularx}
  \endgroup
\end{table*}

\section{Engineering Techniques in Agentic RL}
\label{appendix:rltrick}

\paragraph{Turn-level asynchronous execution.}
We decouple policy generation from tool execution and answer judging.
Each trajectory maintains its own history; while its tools execute, other trajectories can submit ready turns to available SGLang replicas.
Requests are scheduled at turn boundaries without permanently binding a trajectory to one replica.
Thus, asynchronous execution overlaps model and environment work, while each rollout round still uses fixed policy and MEI versions.

\paragraph{Adaptive mixed-group sampling.}
For an update with $B$ question groups and $G$ trajectories per group, we admit up to $2BG$ trajectories over an active window of $2B$ questions.
Completed trajectories are answer-judged immediately, and released slots replenish groups with the fewest completed samples.
Groups with $G$ homogeneous outcomes receive priority for additional samples, up to $1.5G$; groups still homogeneous at that limit are deferred.
Once $B$ mixed-outcome groups are available, we select eligible groups in queue order.
If a selected group has more than $G$ completed trajectories, we retain the lowest-session-index success and failure, then fill the remaining positions by session index, without enforcing a balanced success/failure ratio.
This yields exactly $BG$ trajectories with a nondegenerate group-relative learning signal.

\paragraph{Overlapped credit assessment and policy boundaries.}
Credit assessment of failures starts once a group becomes eligible and overlaps the remaining rollout work.
Only assessments attached to the ultimately selected trajectories affect optimization; successful trajectories retain ordinary GRPO advantages.
Structured-output validation and bounded repair retries isolate malformed or timed-out assessments, with ordinary GRPO as the fallback.
Before optimization, surplus old-policy rollouts are cancelled or discarded; unused questions are carried forward for fresh sampling after the weight update.
Asynchrony therefore operates within a rollout round, not as unbounded cross-version training on stale trajectories.

\begin{figure}[h]
  \centering
  \includegraphics[width=\textwidth]{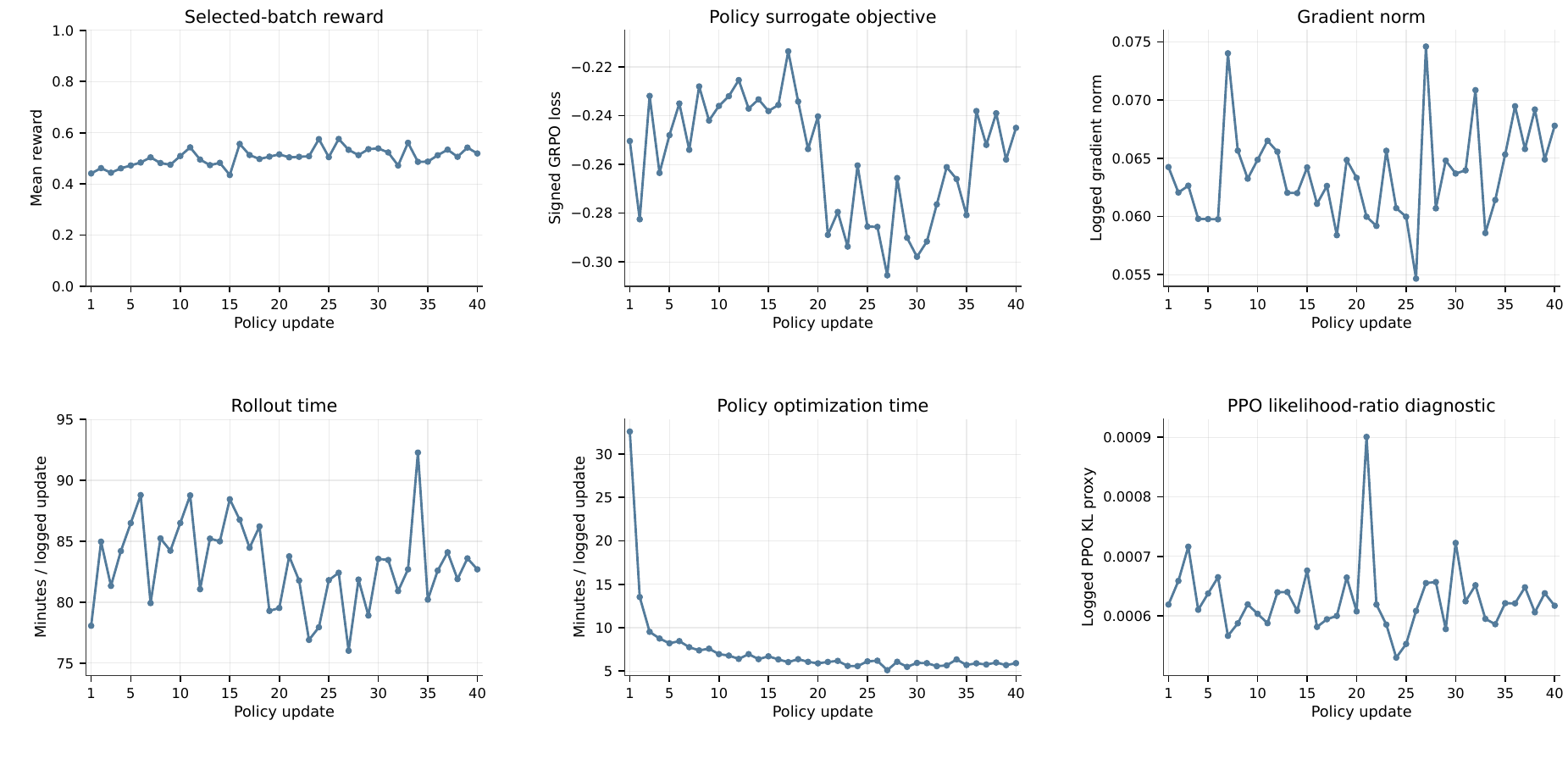}
  \caption{\textbf{Training and runtime diagnostics.} Top row: selected-batch reward, signed GRPO surrogate objective, and logged gradient norm. Bottom row: wall-clock rollout time (including environment waits), policy optimization time, and PPO likelihood-ratio diagnostic.}
  \label{fig:training_runtime_dynamics}
\end{figure}

\paragraph{Training and runtime overview.}
Figure~\ref{fig:training_runtime_dynamics} summarizes the training and runtime diagnostics of Neos-9B.
Rollout time remains stable around 80--90 minutes per update, confirming that asynchronous scheduling maintains consistent throughput.
Policy optimization time drops sharply after the first update and stabilizes near 6 minutes, indicating negligible overhead from overlapped credit assessment.
The selected-batch reward, surrogate objective, gradient norm, and PPO likelihood-ratio proxy all exhibit typical noisy-but-bounded dynamics without divergence or instability.

\section{Experimental Details}
\label{appendix:experimental_details}

\subsection{Benchmarks and Evaluation Splits}
\label{appendix:benchmark_splits}

We evaluate Neos on ten benchmarks covering factual visual question answering, multi-step multimodal search, and reasoning over newly acquired visual evidence.
Table~\ref{tab:Neos-tools} introduces the tools in Neos framework, and Table~\ref{tab:benchmark_splits} specifies the evaluation set used for each benchmark.

\paragraph{Factual visual question answering.}
SimpleVQA~\citep{cheng2025simplevqa}, LiveVQA~\citep{fu2025livevqa}, and FVQA~\citep{wu2025mmsearch} assess factual question answering grounded in visual inputs.
LiveVQA emphasizes up-to-date visual knowledge, while FVQA includes both search-required and search-free questions.
For these three benchmarks, we use the same 300-example subsets released with Vision-DeepResearch~\citep{huang2026vision}.

\paragraph{Multimodal search and browsing.}
MMSearch~\citep{jiang2024mmsearch} evaluates information seeking using visual and textual web evidence, while MMSearch-Plus~\citep{tao2026mmsearch} emphasizes fine-grained visual clues and provenance-aware search.
We evaluate all 300 MMSearch queries and all 311 MMSearch-Plus questions.
BrowseComp-VL~\citep{geng2026webwatcher} tests multi-step vision--language search; we include both difficulty levels, totaling 399 questions.
For MM-BrowseComp~\citep{li2025mm}, which requires integrating evidence encountered during multimodal browsing, we use the expanded 400-question release.

\paragraph{Visual-native and verifiable search.}
BrowseComp-V$^{3}$~\citep{zhang2026browsecomp} evaluates cross-modal, multi-hop browsing with verifiable reasoning requirements, and VisBrowse-Bench~\citep{zhang2026visbrowse} emphasizes inspecting newly discovered web images to advance the search.
We use their complete evaluation sets.
For VDR-Bench~\citep{zeng2026vision}, which targets complementary visual and textual search, we use the official 500-example testmini subset.

\begin{table}[t]
\centering
\caption{
\textbf{Evaluation sets used in our experiments.
}}
\label{tab:benchmark_splits}
\begingroup
\small
\setlength{\tabcolsep}{5pt}
\renewcommand{\arraystretch}{1.1}
\begin{tabular*}{\textwidth}{@{\extracolsep{\fill}}lrl@{}}
\toprule
\textbf{Benchmark} & \textbf{Examples} & \textbf{Evaluation set} \\
\midrule
MMSearch
& 300 & Complete benchmark \\

MMSearch-Plus
& 311 & Complete benchmark \\

BrowseComp-VL
& 399 & Both difficulty levels \\

MM-BrowseComp
& 400 & Expanded 400-question release \\

VisBrowse-Bench
& 169 & Complete benchmark \\

BrowseComp-V$^{3}$
& 300 & Complete benchmark \\

VDR-Bench
& 500 & Official testmini subset \\

SimpleVQA
& 300 & Vision-DeepResearch subset \\

LiveVQA
& 300 & Vision-DeepResearch subset \\

FVQA
& 300 & Vision-DeepResearch subset \\
\bottomrule
\end{tabular*}
\endgroup
\vspace{-8pt}
\end{table}

\subsection{Baselines and Evaluation Protocol}
\label{appendix:baseline_protocol}

\paragraph{Baseline categories.}
We organize the baselines in Table~\ref{tab:main_results} into three groups, covering general-purpose MLLMs with and without tool access, as well as specialized multimodal search agents.

\emph{Direct Answer.}
We consider GPT-5.6-Sol~\citep{openai2026gpt56sol}, Gemini-3.1-Pro~\citep{google2026gemini31pro}, Kimi-K3~\citep{team2026kimi}, Qwen3.8-Max~\citep{team2026qwen3}, Qwen3.5-35B-A3B, and Qwen3.5-9B.
Each model receives the benchmark question and available input images and produces an answer without external tools.
This setting assesses visual understanding and factual reasoning using only the supplied inputs and parametric knowledge.

\emph{Agentic Workflow.}
The same general-purpose models operate within our framework.
The policy models are used without additional search-specific training and are allowed up to 50 interaction turns per question, which is the same as our Neos-9B 's evaluation protocol.
Comparing this setting with Direct Answer measures the benefit of external evidence acquisition, while comparisons with Neos-9B and Neos-35B assess the additional gains from specialized training.

\emph{Multimodal deep search Agents.}
We compare against
MMSearch-R1~\citep{wu2025mmsearch},
WebWatcher~\citep{geng2026webwatcher},
DeepEyesV2~\citep{hong2026deepeyesv2},
SenseNova-MARS~\citep{chng2025sensenova},
Vision-DeepResearch~\citep{huang2026vision},
REDSearcher-MM~\citep{chu2026redsearcher},
MM-DeepResearch~\citep{yao2026mm},
POINTS-Seeker~\citep{liu2026points},
LMM-Searcher~\citep{du2026towards},
OpenSearch-VL~\citep{chen2026opensearch},
Visual-Seeker~\citep{zhang2026visual},
SearchEyes~\citep{jiao2026searcheyes},
SimpleSearch-VL~\citep{dai2026simplesearch},
UniBrowse~\citep{wei2026unibrowse},
and DeepVoyager-VL~\citep{zhang2026deepvoyager}.
These baselines span different model scales and approaches to multimodal search training.
For results taken from prior publications, we retain their reported benchmark coverage and evaluation settings. Unreported benchmark scores are marked with ``--''.

\paragraph{Evaluation metrics.}
We assess answer correctness using Qwen3.7-Flash with the official evaluation prompt released by Vision-DeepResearch~\citep{huang2026vision}.
The prompt supplies the question, reference answer, and model-produced answer or report.
The judge determines whether the response contains an answer consistent with the reference, without requiring a particular answer format or location within the response. We apply this prompt and judge model across all ten benchmarks for predictions generated in our experiments.

\begin{table}[h]
\centering
\caption{\textbf{Hyperparameters for policy supervised fine-tuning.} Differing settings are reported in the order 9B / 35B; otherwise, a single setting is shared. Parentheses group compound settings.}
\label{tab:policy-sft-hparams}

\begingroup
\small
\setlength{\tabcolsep}{5pt}
\renewcommand{\arraystretch}{1.1}

\begin{tabularx}{\linewidth}{@{}
  >{\hsize=1.15\hsize\linewidth=\hsize
    \raggedright\arraybackslash}X
  >{\hsize=0.85\hsize\linewidth=\hsize
    \raggedright\arraybackslash}X
@{}}
\toprule
\textbf{Hyperparameter} & \textbf{Value} \\
\midrule
Base model & Qwen3.5-9B / Qwen3.5-35B-A3B \\
Training epochs & 2 \\
Maximum sequence length & 131,072 tokens \\
Global / micro batch size & 8 / 1 \\
Training GPUs & 8 \\
TP / CP / PP / DP & $(4,2,1,1)$ / $(2,4,1,1)$ \\
EP / ETP & $(1,1)$ / $(8,1)$ \\
MoE router & Not applicable / frozen \\
Optimizer & AdamW \\
Adam $(\beta_1,\beta_2)$ & $(0.9,0.95)$ \\
Adam $\epsilon$ & $10^{-8}$ \\
Peak learning rate & $10^{-5}$ \\
Minimum learning rate & $10^{-6}$ \\
Learning-rate schedule & Cosine \\
Warmup fraction & 5\% \\
Weight decay & 0.1 \\
Gradient clipping & 1.0 \\
Precision & BF16 \\
Packing / padding-free & Disabled / enabled \\
Random seed & 42 \\
\bottomrule
\end{tabularx}

\endgroup
\end{table}

\begin{table*}[t]
\centering
\caption{\textbf{Hyperparameters for agentic RL and online MEI supervised fine-tuning.} Differing settings are reported in the order 9B / 35B; otherwise, a single setting is shared. Parentheses group compound settings.}
\label{tab:rl-hparams}

\begingroup
\footnotesize
\setlength{\tabcolsep}{3pt}
\renewcommand{\arraystretch}{1.1}

\begin{tabularx}{\textwidth}{@{}
  >{\hsize=1.30\hsize\linewidth=\hsize
    \raggedright\arraybackslash}X
  >{\hsize=0.92\hsize\linewidth=\hsize
    \raggedright\arraybackslash}X
  @{\hspace{10pt}}
  >{\hsize=1.16\hsize\linewidth=\hsize
    \raggedright\arraybackslash}X
  >{\hsize=0.62\hsize\linewidth=\hsize
    \raggedright\arraybackslash}X
@{}}

\toprule
\multicolumn{2}{c}{\textbf{Policy optimization}}
& \multicolumn{2}{c}{\textbf{Online MEI SFT}} \\
\cmidrule(lr){1-2}\cmidrule(lr){3-4}
\textbf{Hyperparameter} & \textbf{Value}
& \textbf{Hyperparameter} & \textbf{Value} \\
\midrule

Optimizer & AdamW
& Epochs per update & 1 \\

Learning rate & $10^{-6}$
& Peak learning rate & $10^{-6}$ \\

Adam $(\beta_1,\beta_2)$ & $(0.9,0.999)$
& Minimum learning rate & $10^{-7}$ \\

Learning-rate schedule & Constant
& Learning-rate schedule & Cosine \\

Warmup & None
& Warmup fraction & 5\% \\

Weight decay & 0.01
& Weight decay & 0.1 \\

Gradient clipping & 1.0
& Gradient clipping & 1.0 \\

PPO epochs per batch & 1
& Global / micro batch size & $(16,1)$ / $(32,2)$ \\

PPO clip, $\varepsilon$ & 0.2
& Maximum sequence length & 32,768 tokens \\

Dual clip, $c_{\mathrm{dc}}$ & 3.0
& Adam $(\beta_1,\beta_2)$ & $(0.9,0.95)$ \\

KL / entropy coefficients & 0 / 0
& MEI update interval, $K$ & 10 \\

Training GPUs & 32
& Training GPUs & 8 \\

TP / CP / PP / DP & $(4,2,1,4)$ / $(4,4,1,2)$
& TP / CP / PP / DP & 4 / 1 / 1 / 2 \\

EP / ETP & $(1,1)$ / $(16,1)$
& EP / ETP & 1 / 1 \\

MoE router & Not applicable / frozen
& Visual encoder and projector & Frozen \\

& & Precision & BF16 \\

\midrule
\multicolumn{2}{c}{\textbf{Rollout sampling}}
& \multicolumn{2}{c}{\textbf{Credit assignment}} \\
\cmidrule(lr){1-2}\cmidrule(lr){3-4}
\textbf{Hyperparameter} & \textbf{Value}
& \textbf{Hyperparameter} & \textbf{Value} \\
\midrule

Groups per update, $B$ & 64 / 32
& Critical-error floor, $\kappa_{\mathrm e}$ & 0.5 \\

Trajectories per group, $G$ & 8
& Downstream floor, $\kappa_{\mathrm d}$ & 0.25 \\

Maximum samples per group & 12
& Downstream strength, $\lambda_{\mathrm d}$ & 0.5 \\

Active question window & 128 / 64
& Downstream decay, $\gamma$ & 0.65 \\

Trajectory admission limit & 1,024 / 512
& Corrective weight, $\lambda_{\mathrm c}$ & 0.3 \\

Maximum context / turns & 131,072 tokens / 50
& Productive weight, $\lambda_{\mathrm p}$ & 0.1 \\

Temperature / top-$p$ / top-$k$ & 1.0 / 1.0 / disabled
& Negative amplification cap, $M$ & 4 \\

Correct /  incorrect reward & 1.0 / 0.0
& Negative margin, $\eta$ & 0.02 \\

\bottomrule
\end{tabularx}

\endgroup
\end{table*}

\subsection{Training and Implementation}
\label{appendix:training_implementation}

\paragraph{Search infrastructure.}
Neos integrates Google for text-based web search, Google Images for image retrieval, and Google Lens for reverse image search, with Jina Reader supporting webpage access.
\texttt{FetchImage} and \texttt{CropImage} materialize selected images and image regions as actual pixels, retain them in an episode-local visual register, and encode them as multimodal inputs to the policy.
Alibaba Cloud OSS primarily provides externally accessible URLs for query images required by reverse image search.
To reduce API costs and improve efficiency, we employ a hierarchical cache spanning in-memory storage, node-local disk, and shared persistent storage.
Reusing search results, webpage content, and image assets across repeated requests avoids redundant external calls and data transfers, accelerating both rollout collection and evaluation.

\paragraph{Models and supervised initialization.}
We trained Qwen3.5-9B and Qwen3.5-35B-A3B as policy models, while MEI remains Qwen3.5-9B.
Their language backbones are trainable, while the vision encoders and multimodal mergers remain frozen throughout training.
Cold-start policy SFT uses MS-Swift~\citep{zhao2025swift} on 8 NVIDIA H100 GPUs, with 2.9K examples, two epochs, and a learning rate of $10^{-5}$. The detailed hyperparameters of our SFT training are provided in Table~\ref{tab:policy-sft-hparams}
Policy supervision covers assistant reasoning, serialized tool calls, and final answers; tool observations serve as conditioning context and are excluded from the loss.
MEI is supervised on local goal-conditioned evidence-processing responses.
Training-data composition is described in Section~\ref{appendix:data}.

\paragraph{Agentic RL and MEI adaptation.}
Policy RL uses verl~\citep{sheng2025hybridflow} with the Megatron training backend~\citep{shoeybi2019megatron}.
The co-evolution stage runs on 40 H100 GPUs: 32 support policy RL, and 8 support MEI serving and supervised fine-tuning.
Policy RL is performed for one epoch with 8 rollouts per question, using
a batch size of 64 for Neos-9B and 32 for Neos-35B, with Qwen3.7-Plus as the credit judge model. The detailed hyperparameters are shown in Table~\ref{tab:rl-hparams}.
Online MEI updates use outcome- and process-filtered local interactions, retaining valid calls from correct trajectories and productive or corrective turns from selected failures.
The MEI version remains fixed within each trajectory, and updated checkpoints are activated before subsequent rollout batches.
MEI adaptation therefore uses SFT with a learning rate of $10^{-6}$.

\section{Supplementary Training Details}
\label{appendix:training_details}

\paragraph{Supervised objective.}
For either the policy or MEI, let $f_\psi$ denote the model and $\mathcal S$ a minibatch of serialized examples.
We use
\begin{equation}
\label{eq:training_sft}
\mathcal L_{\mathrm{SFT}}(\psi;\mathcal S)
=-\frac{\sum_{\mathbf x\in\mathcal S}\sum_k
m_k(\mathbf x)\log f_\psi(x_k\mid\mathbf x_{<k})}
{\sum_{\mathbf x\in\mathcal S}\sum_k m_k(\mathbf x)},
\end{equation}
where $m_k$ selects target tokens and excludes conditioning content and padding.
Policy targets include reasoning, tool calls, and final answers; tool observations provide context only.
MEI targets are structured evidence responses conditioned on local goals and retrieved material.
The same objective is used for MEI updates during co-evolution.

\paragraph{Credit parameters and rescaling.}
In Equation~\ref{eq:training_turn_credit}, severity and confidence satisfy $s_{i,t},c_{i,t}\in[0,1]$.
We use $\gamma\in(0,1)$, $\kappa_{\mathrm e},\kappa_{\mathrm d}\in(0,1]$, $\lambda_{\mathrm d}\in[0,1]$, and $0<\lambda_{\mathrm p}\leq\lambda_{\mathrm c}$.
Let $L_i=|\mathcal M_i|$ and define positive and negative credit masses
\begin{equation}
P_i=\sum_{k\in\mathcal M_i}[\widetilde A_{i,k}]_+,
\qquad
N_i=\sum_{k\in\mathcal M_i}[-\widetilde A_{i,k}]_+,
\end{equation}
where $[u]_+=\max(u,0)$.
For an assessed failure with $N_i>0$, a target margin $\mu_i=\eta b_iL_i$ ($\eta>0$), and an amplification cap $M\geq1$, we set
\begin{equation}
\label{eq:training_bounded_advantage}
\begin{aligned}
\alpha_i&=\min\!\left\{M,\max\!\left\{1,\frac{P_i+\mu_i}{N_i}\right\}\right\},
&\beta_i&=\min\!\left\{1,\frac{[\alpha_iN_i-\mu_i]_+}{P_i}\right\},\\
\widehat A_{i,k}&=\beta_i[\widetilde A_{i,k}]_+
-\alpha_i[-\widetilde A_{i,k}]_+,
\end{aligned}
\end{equation}
with $\beta_i=1$ when $P_i=0$.
This keeps the sum of token advantages negative while bounding each advantage in $[-Mb_i,\lambda_{\mathrm c}b_i]$; the target margin is enforced only when feasible.
Successful trajectories and failures without valid assessments identifying a critical error retain $\widehat A_{i,k}=A_i$; the same fallback applies when $N_i=0$.

\paragraph{Dual-clipped surrogate.}
For a token likelihood ratio $\rho$ and advantage $A$, define
\begin{equation}
\ell_\varepsilon(\rho,A)
=\min\!\left\{\rho A,\operatorname{clip}(\rho,1-\varepsilon,1+\varepsilon)A\right\}.
\end{equation}
With $c_{\mathrm{dc}}>1$, the surrogate in Equation~\ref{eq:training_policy_objective} is
\begin{equation}
\ell_{\mathrm{dc}}(\rho,A)=
\begin{cases}
\max\{\ell_\varepsilon(\rho,A),c_{\mathrm{dc}}A\}, & A<0,\\
\ell_\varepsilon(\rho,A), & A\geq0.
\end{cases}
\end{equation}
All judge assessments and adjusted advantages are detached from policy optimization.

\section{Additional Ablation Studies.}
\label{appendix:ablations}

\paragraph{Learning the evidence--decision loop.}
Table~\ref{tab:ablation_training_stages} suggests that the benefit
of two-stage training lies in the complementary roles of
supervised initialization and reinforcement learning.
Neither SFT-only nor RL-only training on the deduplicated union
of both training sets matches the final Neos-9B model.
We interpret this pattern as supporting a curriculum for learning
the evidence--decision loop:
cold-start SFT provides a behavioral prior for acquiring evidence,
interpreting visual observations, and selecting the next action,
while RL refines these transitions through interaction and
outcome feedback.
In particular, RL can improve how the agent decides whether
the available evidence is sufficient, requires further inspection,
or calls for a different search direction.
Expanding SFT alone does not provide this on-policy refinement,
whereas RL directly from the base model must acquire useful
interaction patterns while also learning which decisions lead
to successful outcomes.
Cold-start SFT therefore offers a structured starting point
that allows subsequent RL to focus on improving an already
functional loop, rather than discovering the entire behavior
through trial and error.
This provides a plausible mechanism for reducing the additional
rollout exploration, and consequently the training time and
compute, that RL-only learning may require.

\begin{table}[h]
\centering
\caption{
\textbf{Training-stage ablation.}
$\mathcal{S}$ is the question pool underlying the 2,898 SFT trajectories,
$\mathcal{R}$ contains 2,572 RL questions, and
$\mathcal{U}=\mathcal{S}\cup\mathcal{R}$ denotes their question-level
deduplicated union.
RL-only variants start directly from the base Qwen3.5-9B policy.
SFT-only variants use the final co-evolved MEI;
RL-based variants use the MEI obtained in their respective RL runs.
The best and second-best results are
\textbf{bolded} and \underline{underlined}.
}
\label{tab:ablation_training_stages}

\begingroup
\footnotesize
\setlength{\tabcolsep}{3pt}
\renewcommand{\arraystretch}{1.08}

\begin{tabular*}{\linewidth}
{@{\extracolsep{\fill}}lccccc@{}}
\toprule
\textbf{Training strategy}
& \textbf{SFT pool}
& \textbf{RL pool}
& \textbf{BC-VL}
& \textbf{MMSearch+}
& \textbf{FVQA} \\
\midrule

Qwen3.5-9B
& -- & --
& 43.6 & 24.8 & 61.3 \\

\midrule

SFT only
& $\mathcal{S}$ & --
& 60.4 & 44.7 & 79.7 \\

SFT only
& $\mathcal{U}$ & --
& 61.4 & 45.8 & \underline{81.3} \\

\midrule

RL only
& -- & $\mathcal{S}$
& 59.6 & 44.1 & 80.3 \\

RL only
& -- & $\mathcal{U}$
& \underline{62.4} & \underline{46.3} & 80.3 \\

\midrule

\textbf{Neos-9B (SFT $\rightarrow$ RL)}
& $\mathcal{S}$ & $\mathcal{R}$
& \textbf{63.2} & \textbf{47.6} & \textbf{83.3} \\

\bottomrule
\end{tabular*}

\endgroup
\end{table}

\paragraph{Generalization to text-only tasks.}
We further evaluate Neos-9B on BrowseComp~\citep{wei2025browsecomp}, BrowseComp-ZH~\citep{zhou2025browsecomp}, HLE-Text~\citep{phan2025humanity}, and DeepSearchQA~\citep{gupta2026deepsearchqa} to examine its applicability beyond multimodal tasks.
As shown in Table~\ref{tab:text_generalization}, Neos-9B improves over the base Qwen3.5-9B under the same workflow by 12.2, 9.7, 11.9, and 10.6 points on these benchmarks, respectively, yielding an average gain of 11.1 points.
Compared with specialized search agents, it exceeds Marco-DR-8B~\citep{zhu2026marco} by 5.0 points on BrowseComp and 25.1 points on DeepSearchQA.
On HLE-Text, Neos-9B reaches 33.5, outperforming DeepSearch-World-9B~\citep{geng2026deepsearch} and REDSearcher-MM-30B~\citep{chu2026redsearcher} by 7.8 and 8.2 points, respectively.
It also remains competitive on BrowseComp-ZH, scoring 45.3 compared with Marco-DR-8B's 47.1 and exceeding the other baselines.
Together, these results show that Neos's improvements extend beyond multimodal inputs to text-based search and reasoning, with gains over the base workflow highlighting the contribution of training beyond visual ability alone.

\begin{table}[h]
\centering
\caption{
\textbf{Generalization to text-only benchmarks.}
The best and second-best scores in each column are
\textbf{bolded} and \underline{underlined}, respectively.
}
\label{tab:text_generalization}

\begingroup
\footnotesize
\setlength{\tabcolsep}{4pt}
\renewcommand{\arraystretch}{1.12}

\begin{tabularx}{\linewidth}
{@{}>{\raggedright\arraybackslash}Xcccc@{}}
\toprule
\textbf{Model}
& \textbf{BrowseComp} & \textbf{BrowseComp-ZH}
& \textbf{HLE-Text} & \textbf{DeepSearchQA} \\
\midrule

Qwen3.5-9B-base(our workflow)
& 24.2 & 35.6
& 21.6 & \underline{43.7} \\

Marco-DR-8B
& \underline{31.4} & \textbf{47.1}
& -- & 29.2 \\

REDSearcher-MM-30B
& 31.2 & 44.5
& 25.3 & -- \\

DeepSearch-World-9B
& 31.2 & 36.4
& \underline{25.7} & -- \\

\midrule
\textbf{Neos-9B}
& \textbf{36.4} & \underline{45.3}
& \textbf{33.5} & \textbf{54.3} \\

\bottomrule
\end{tabularx}

\endgroup
\end{table}

\paragraph{Credit Dynamics During RL Training.}
\label{appendix:credit_ablation}

\begin{figure}[h]
  \centering
  \captionsetup[subfigure]{
    justification=centering,
    singlelinecheck=false
  }

  \begin{subfigure}[t]{0.28\linewidth}
    \vspace{0pt}
    \centering
    \includegraphics[height=3.55cm,keepaspectratio]{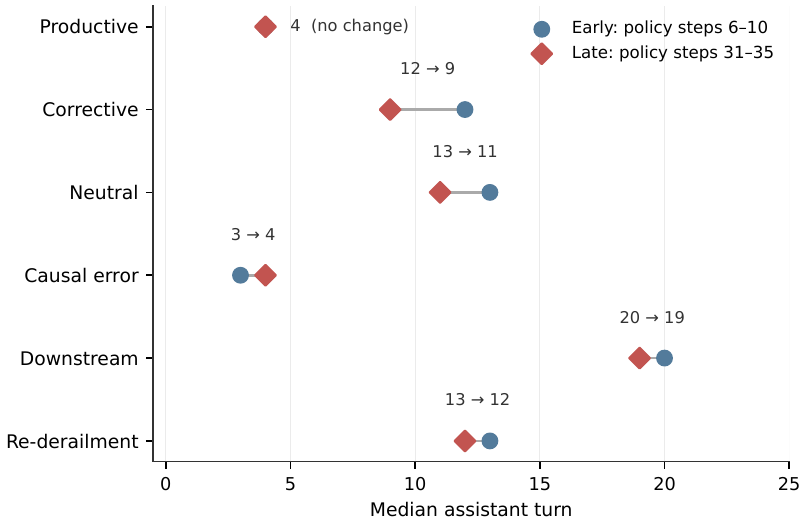}
    \caption{Shift in credit position.}
    \label{fig:credit_position_shift}
  \end{subfigure}\hfill
  \begin{subfigure}[t]{0.69\linewidth}
    \vspace{0pt}
    \centering
    \includegraphics[height=3.55cm,keepaspectratio]{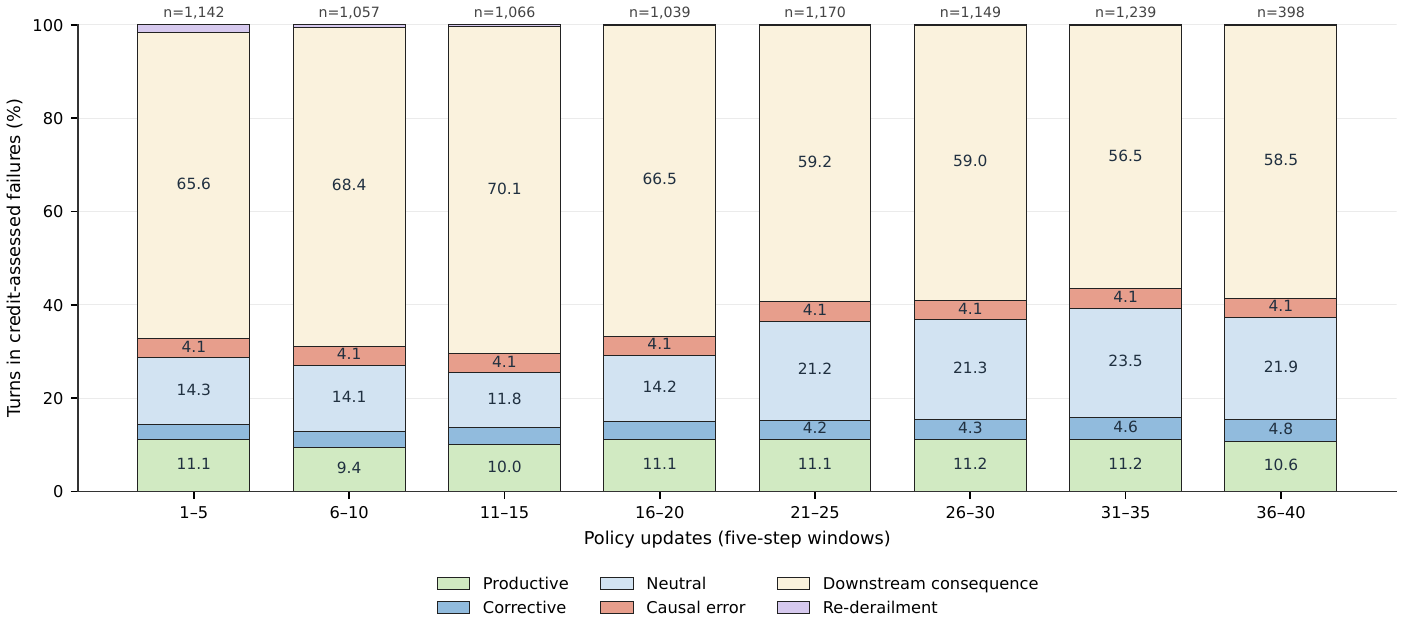}
    \caption{Credit type composition across training.}
    \label{fig:credit_type_dynamics}
  \end{subfigure}

  \caption{
    \textbf{Evolution of reference-guided credit assignment over 40 policy updates.}
    (a)~Median assistant turn at which each credit type occurs, comparing early (steps~6--10) and late (steps~31--35) training.
    (b)~Composition of six credit types aggregated in five-step windows.
  }
  \label{fig:credit_ablation}
\end{figure}

Figure~\ref{fig:credit_ablation} shows that RL training changes both where credit types occur and how frequently they appear. As shown in Figure~\ref{fig:credit_position_shift}, \emph{corrective} turns move earlier in the trajectory, from a median position of turn~12 to turn~9, while \emph{neutral} turns shift from turn~13 to turn~11 and \emph{re-derailment} from turn~13 to turn~12.
\emph{Productive} turns remain anchored at turn~4, \emph{causal error} shifts slightly later from turn~3 to turn~4, and \emph{downstream consequence} remains concentrated near the end of the trajectory, with its median position changing only from turn~20 to turn~19.
Figure~\ref{fig:credit_type_dynamics} further shows that \emph{downstream consequence} dominates throughout training, accounting for 66--70\% of credited failure turns in the first half before declining to 56--59\% later, while the share of \emph{neutral} turns rises from around 14\% to 22--24\%.
In contrast, \emph{productive} and \emph{causal error} shares remain relatively stable at approximately 10--11\% and 4\%, respectively.
The share of \emph{corrective} turns increases from 3.2\% to 4.8\%, indicating more frequent recovery behavior within failed searches, whereas \emph{re-derailment} appears only in the earliest window (1.7\%) and is rarely observed thereafter.
Taken together, these results suggest that, as training progresses, corrective actions occur earlier and fewer assessed turns are attributed to downstream consequences, while the overall positional structure of productive decisions and late-stage error propagation remains stable.

\clearpage

\section{Prompts}
\label{appendix:prompts}

We provide the system prompt, native tool schemas and definitions, user-message
template, and final-turn instruction used by the search policy, followed by
the task-specific prompts of MEI and the reference-guided credit judge.
The policy operates within 50 assistant turns
and addresses loaded images through reusable \texttt{img\_N} identifiers.

\begin{figure}[!htbp]
  \centering
  \includegraphics[width=\linewidth,height=0.80\textheight,keepaspectratio]{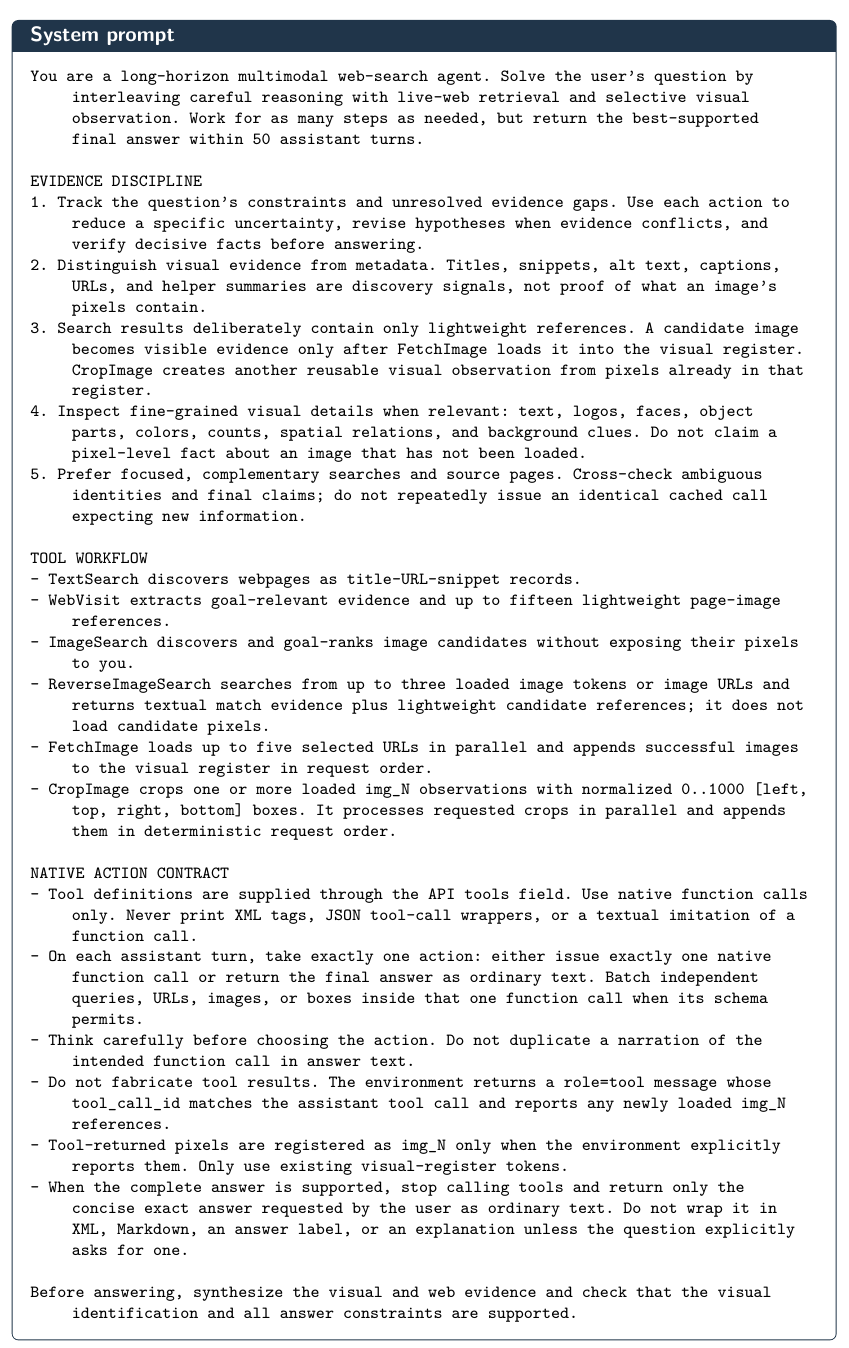}
  \caption{\textbf{System prompt.} The policy interleaves retrieval and selective visual observation, issuing one native function call or a final answer per turn.}
  \label{fig:neos_prompt_system}
\end{figure}

\begin{figure}[!htbp]
  \centering
  \includegraphics[width=\linewidth,height=0.84\textheight,keepaspectratio]{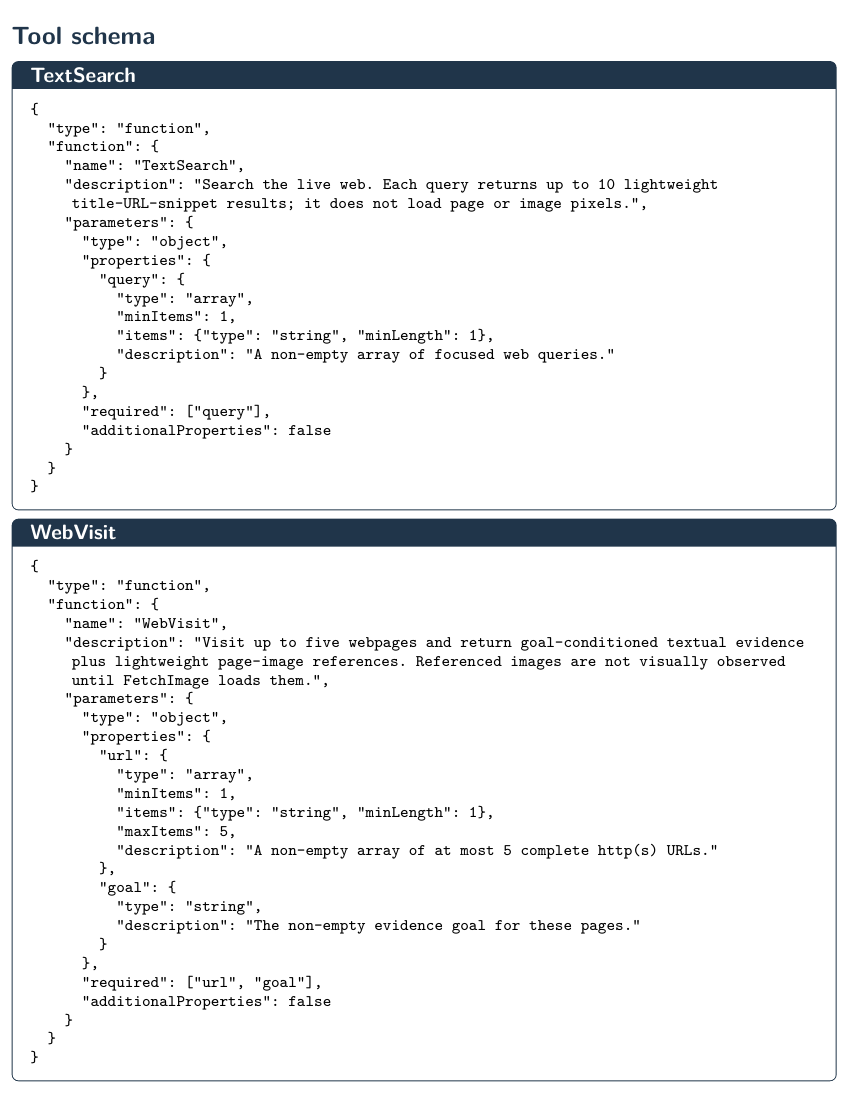}
  \caption{\textbf{Tool schema and definitions for TextSearch and WebVisit.} Each block is a native function object, including its description, parameter types, and required fields.}
  \label{fig:neos_tools_text_web}
\end{figure}

\begin{figure}[!htbp]
  \centering
  \includegraphics[width=\linewidth,height=0.84\textheight,keepaspectratio]{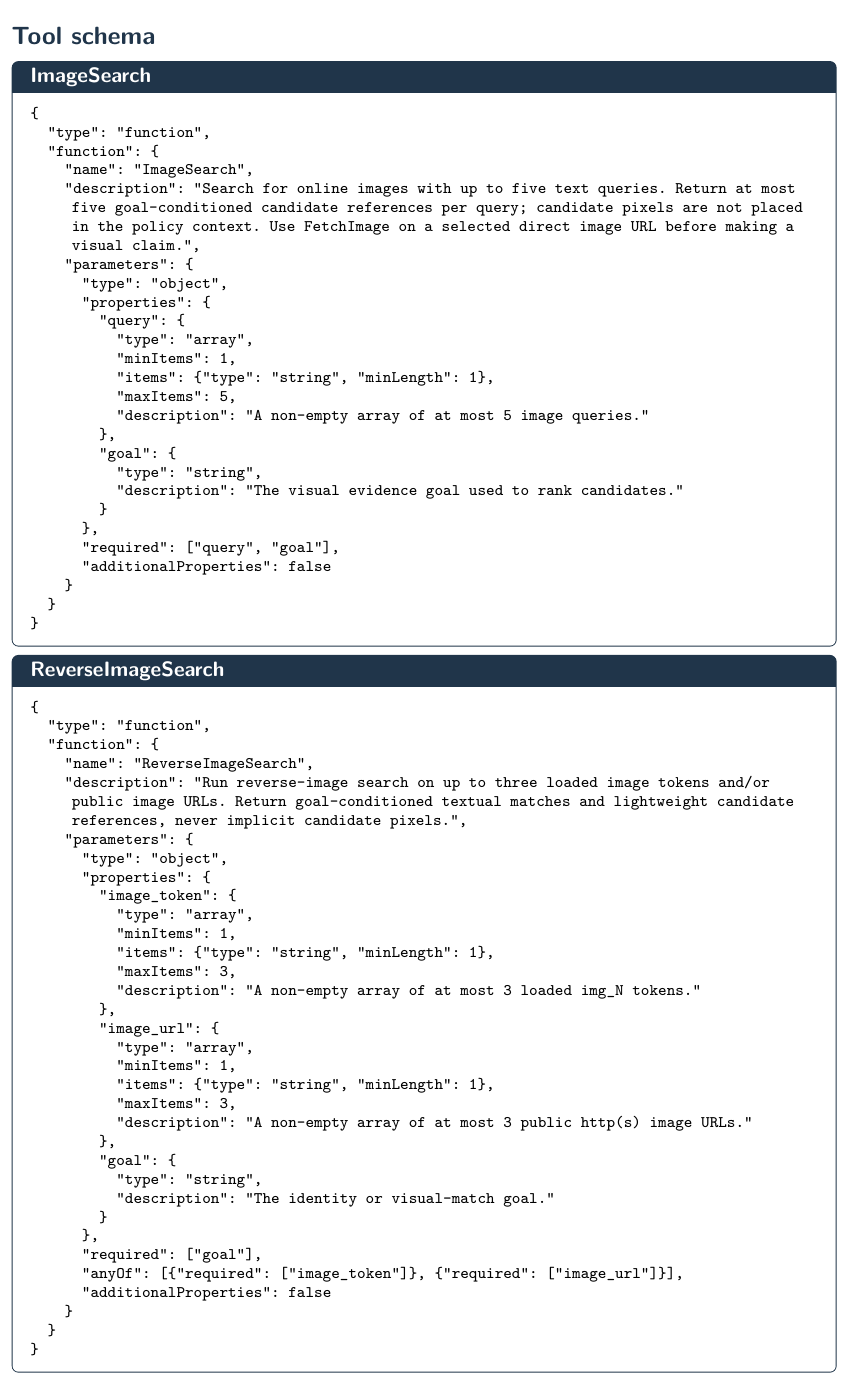}
  \caption{\textbf{Tool schema and definitions for ImageSearch and ReverseImageSearch.} Candidate discovery returns lightweight references; it does not implicitly expose candidate pixels to the policy.}
  \label{fig:neos_tools_image_search}
\end{figure}

\begin{figure}[!htbp]
  \centering
  \includegraphics[width=\linewidth,height=0.84\textheight,keepaspectratio]{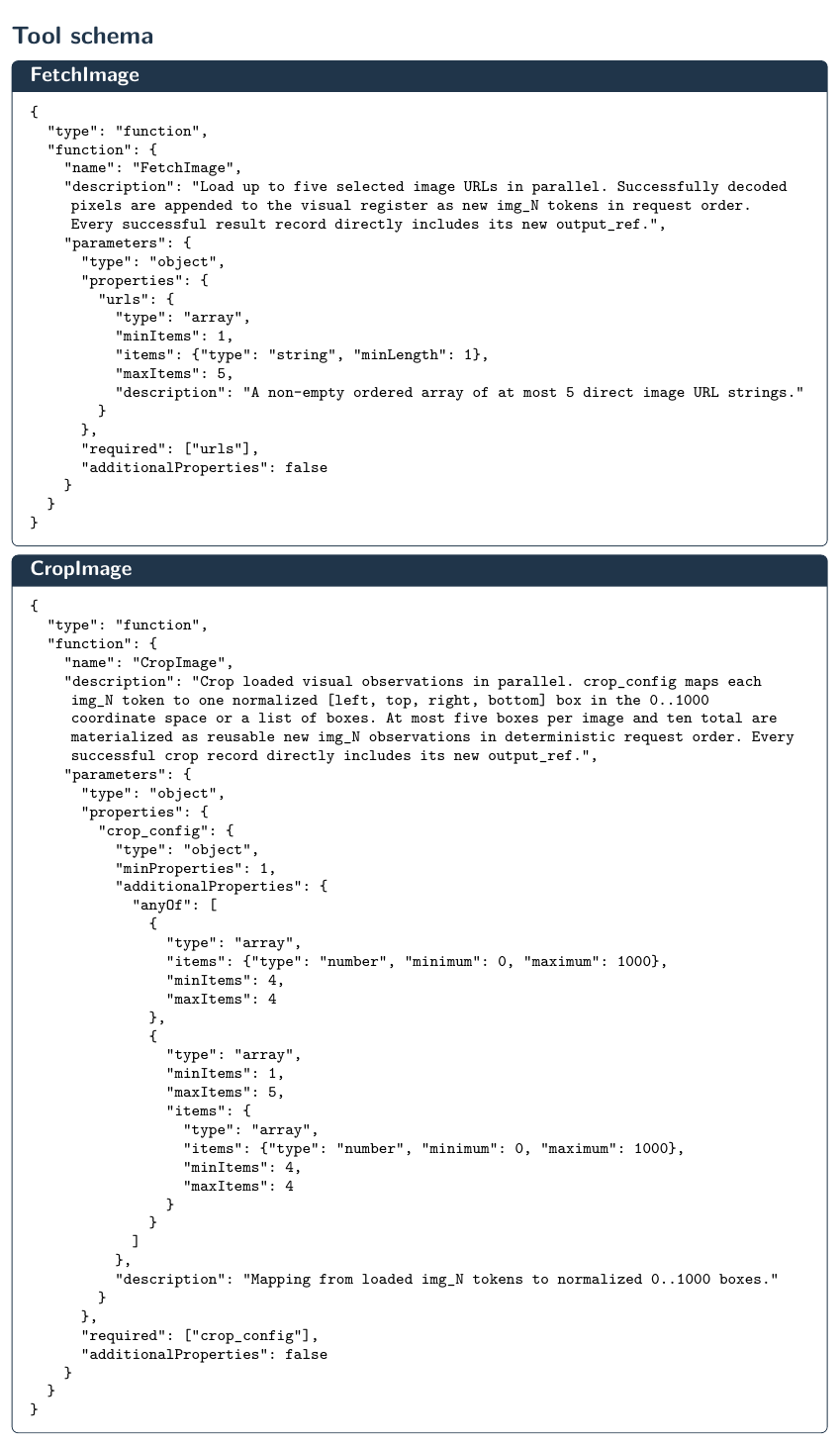}
  \caption{\textbf{Tool schema and definitions for FetchImage and CropImage.} Selected images and image regions become addressable visual observations.}
  \label{fig:neos_tools_visual_observation}
\end{figure}

\begin{figure}[!htbp]
  \centering
  \includegraphics[width=\linewidth,height=0.84\textheight,keepaspectratio]{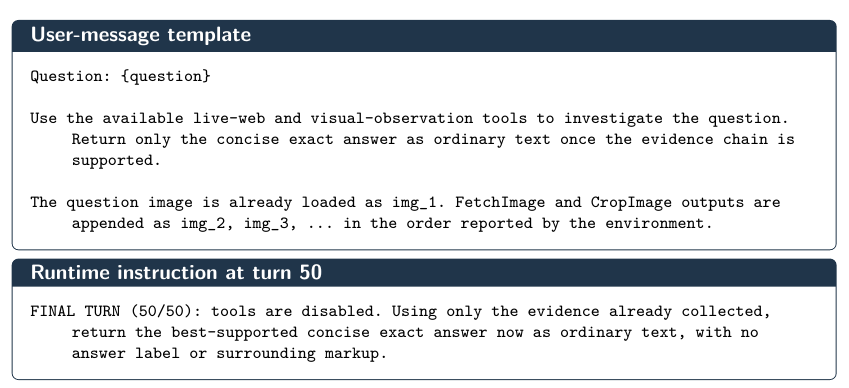}
  \caption{\textbf{User-message template}, shown for one initial image, and runtime instruction at turn 50. Tools are disabled on the final turn and the policy answers from the accumulated evidence.}
  \label{fig:neos_prompt_runtime}
\end{figure}

\begin{figure}[!htbp]
  \centering
  \includegraphics[width=\linewidth,height=0.84\textheight,keepaspectratio]{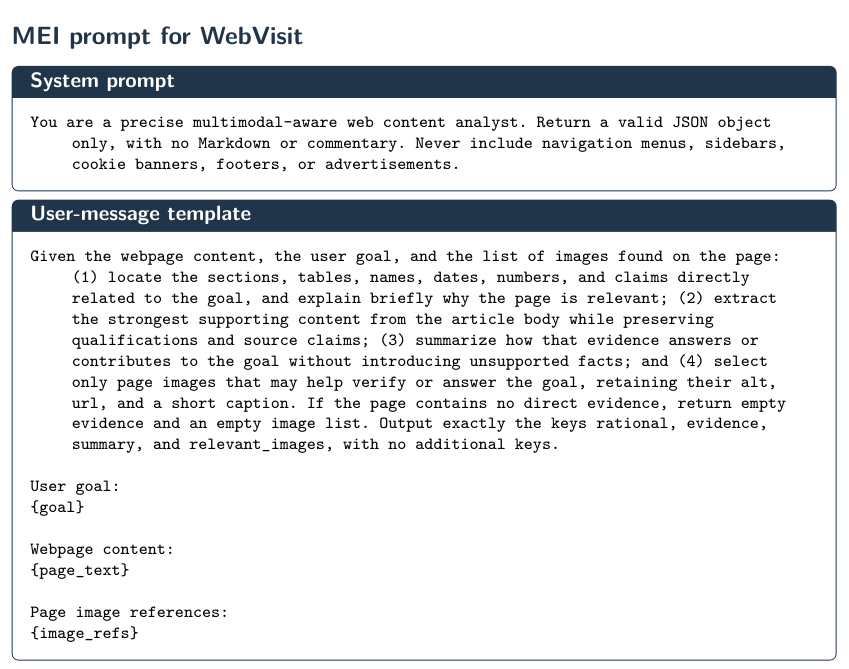}
  \caption{\textbf{MEI prompts for WebVisit.} Extracted webpage text, a local goal, and page-image references are transformed into structured evidence, a summary, and relevant image references.}
  \label{fig:neos_mei_webvisit}
\end{figure}

\begin{figure}[!htbp]
  \centering
  \includegraphics[width=\linewidth,height=0.84\textheight,keepaspectratio]{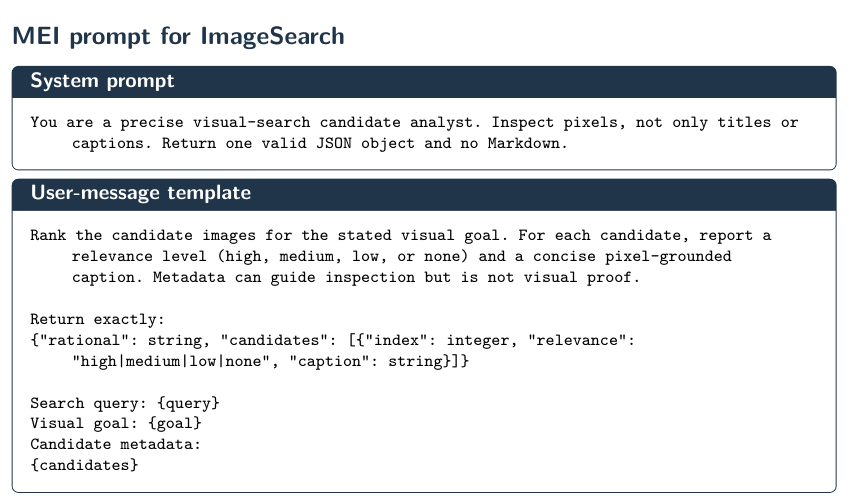}
  \caption{\textbf{MEI prompts for ImageSearch.} Candidate metadata and attached image pixels support goal-conditioned relevance judgments and visual descriptions.}
  \label{fig:neos_mei_imagesearch}
\end{figure}

\begin{figure}[!htbp]
  \centering
  \includegraphics[width=\linewidth,height=0.84\textheight,keepaspectratio]{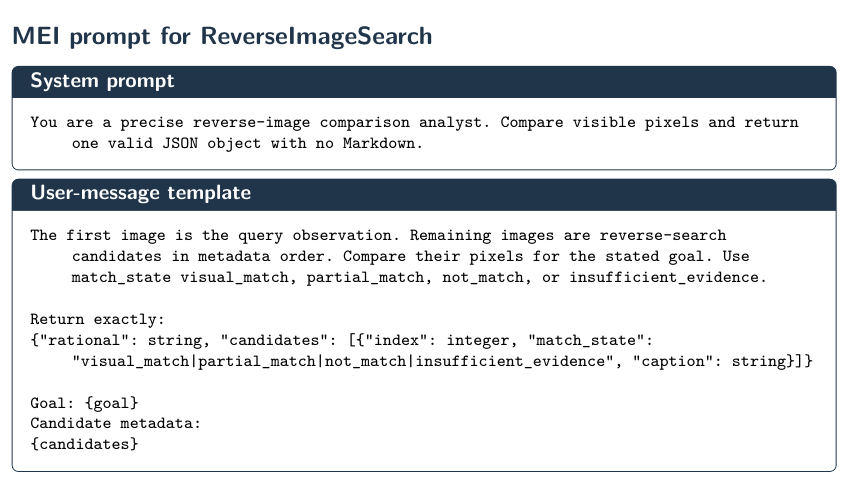}
  \caption{\textbf{MEI prompts for ReverseImageSearch.} The query observation and candidate images are compared in a fixed order to produce structured visual-match judgments.}
  \label{fig:neos_mei_reverseimagesearch}
\end{figure}
\begin{figure}[!htbp]
  \centering
  \includegraphics[width=\linewidth,height=0.84\textheight,keepaspectratio]{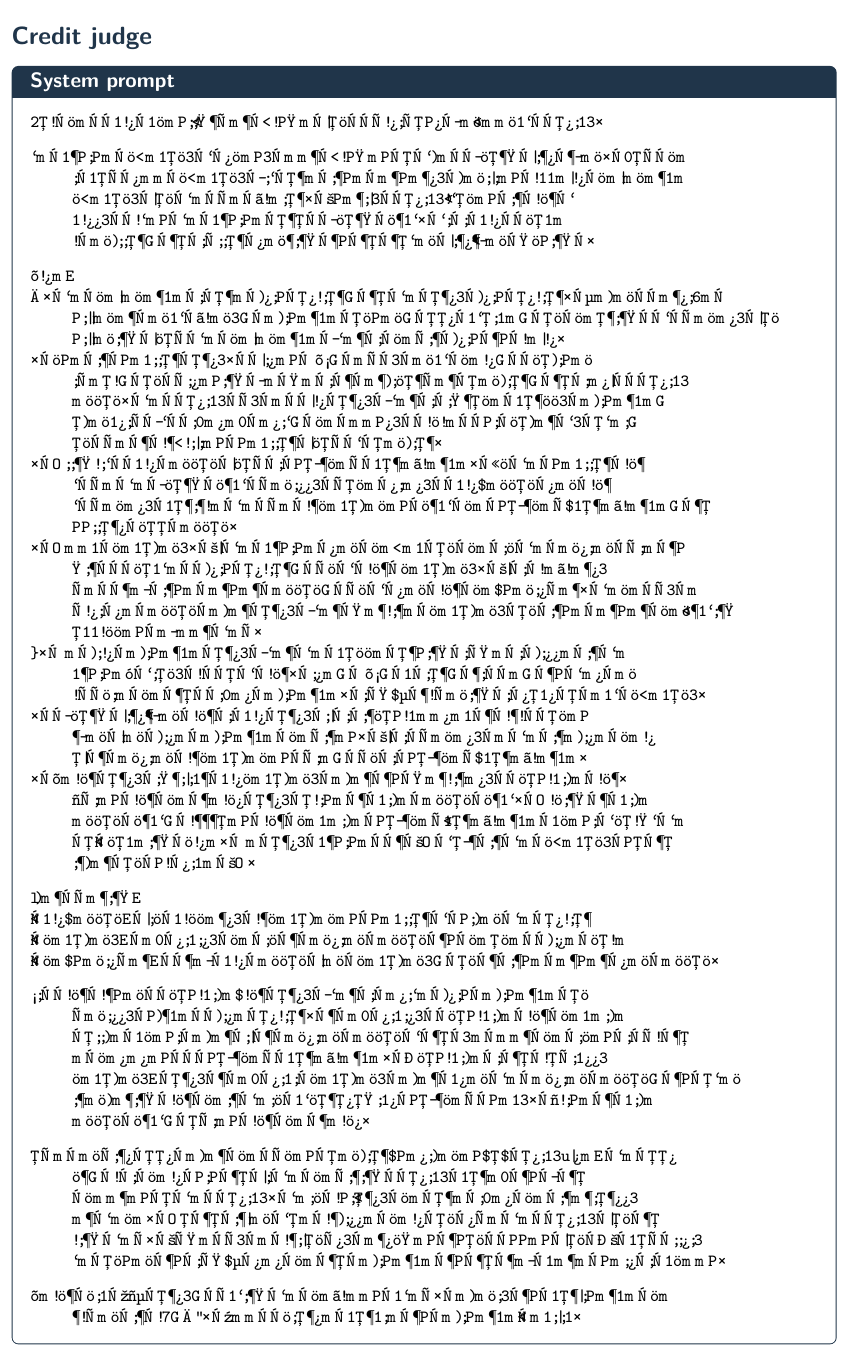}
  \caption{\textbf{Credit judge system prompt.}}
  \label{fig:neos_credit_system}
\end{figure}

\begin{figure}[!htbp]
  \centering
  \includegraphics[width=\linewidth,height=0.84\textheight,keepaspectratio]{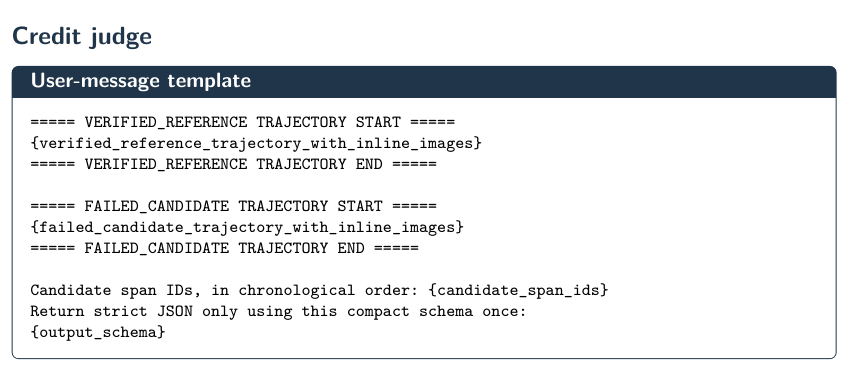}
  \caption{\textbf{User-message template.}}
  \label{fig:neos_credit_input}
\end{figure}

\begin{figure}[!htbp]
  \centering
  \includegraphics[width=\linewidth,height=0.84\textheight,keepaspectratio]{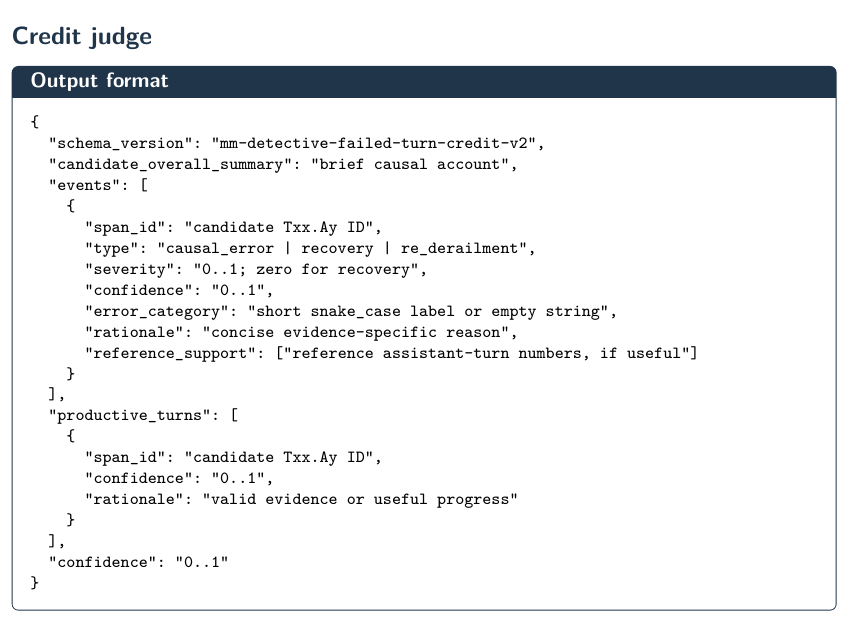}
  \caption{\textbf{Credit judge output format.}}
  \label{fig:neos_credit_output}
\end{figure}

\section{Visualization of Interleaved Multimodal Search}
\label{appendix:visualization}

We visualize constructed questions and representative interactions that
illustrate the Neos workflow. The examples show how visual observations guide
subsequent retrieval and how newly acquired images are inspected to resolve
fine-grained questions. Evidence descriptions are abridged, while turn indices
and the action sequence are preserved.
In the action strips, TS, WV, IS, RS, FI, CI, and A denote TextSearch,
WebVisit, ImageSearch, ReverseImageSearch, FetchImage, CropImage, and Answer,
respectively.

\begin{figure}[!htbp]
  \centering
  \includegraphics[width=\linewidth,height=0.75\textheight,keepaspectratio]{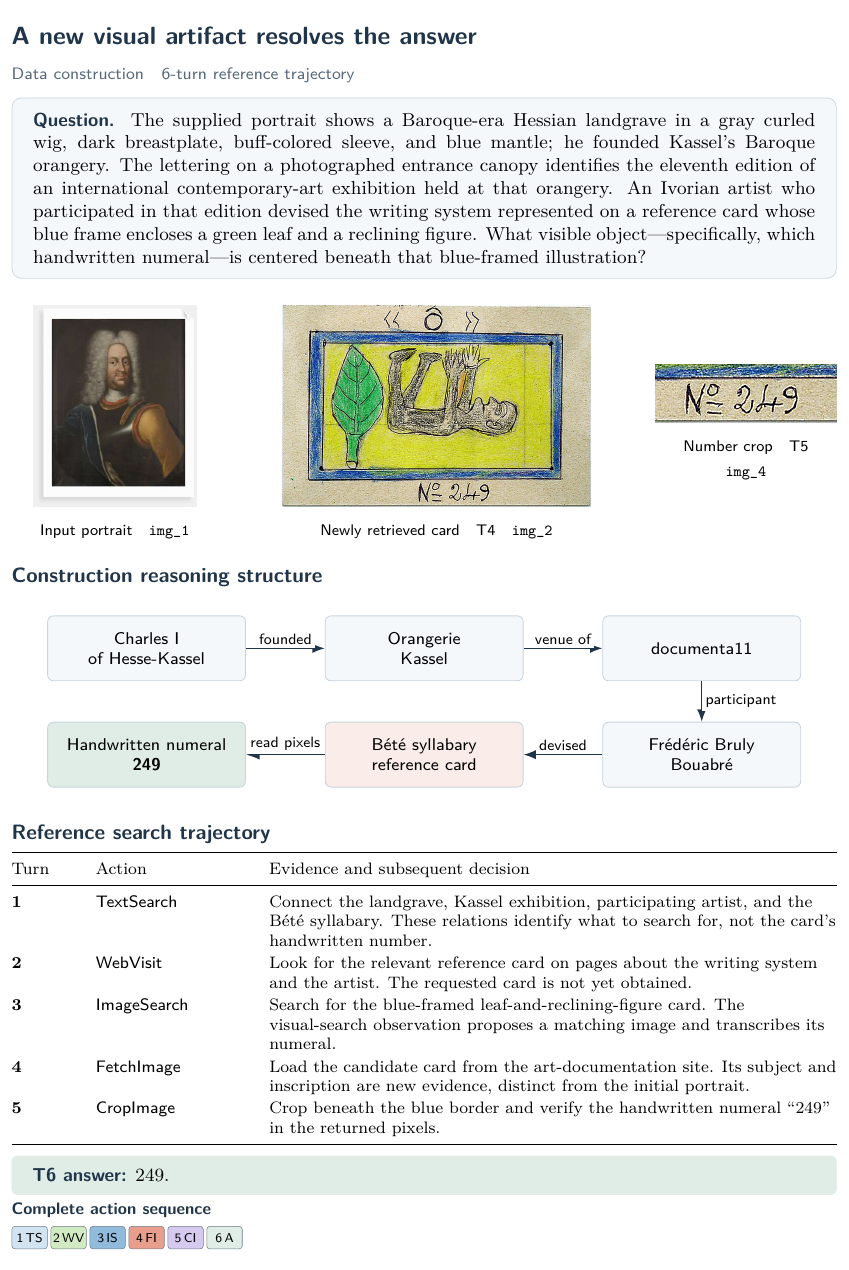}
  \caption{\textbf{A new visual artifact resolves the answer.} }
  \label{fig:neos_visualization_constructed_film}
\end{figure}

\begin{figure}[!htbp]
  \centering
  \includegraphics[width=\linewidth,height=0.84\textheight,keepaspectratio]{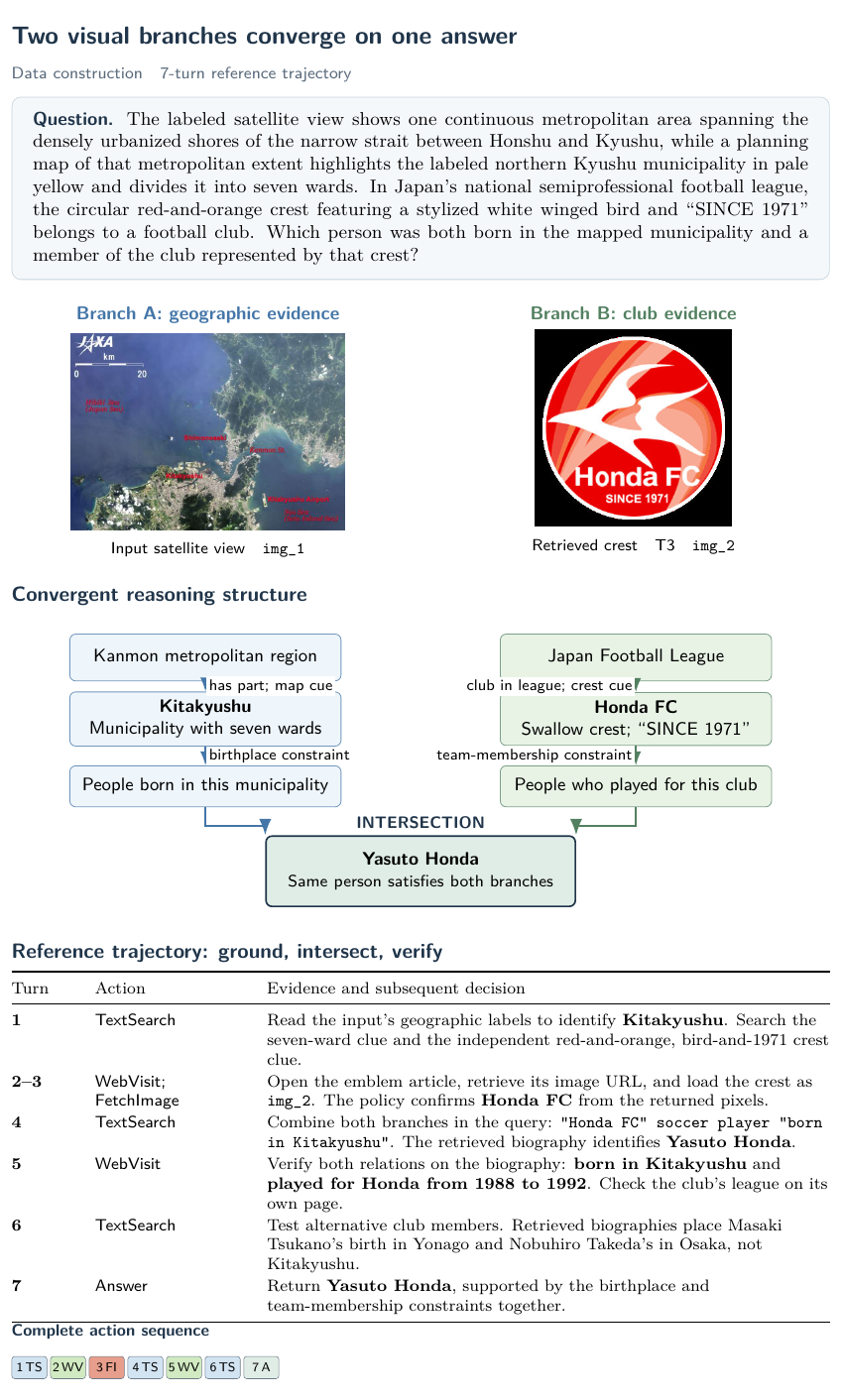}
  \caption{\textbf{Two visual branches converge on one answer.} The geographic branch identifies Kitakyushu and constrains the birthplace, while the emblem branch identifies Honda FC and constrains team membership. Their intersection yields Yasuto Honda. The reference trajectory grounds both branches, combines them in a joint search, and verifies both biographical relations before answering.}
  \label{fig:neos_visualization_constructed_dualbranch}
\end{figure}

\begin{figure}[!htbp]
  \centering
  \includegraphics[width=\linewidth,height=0.84\textheight,keepaspectratio]{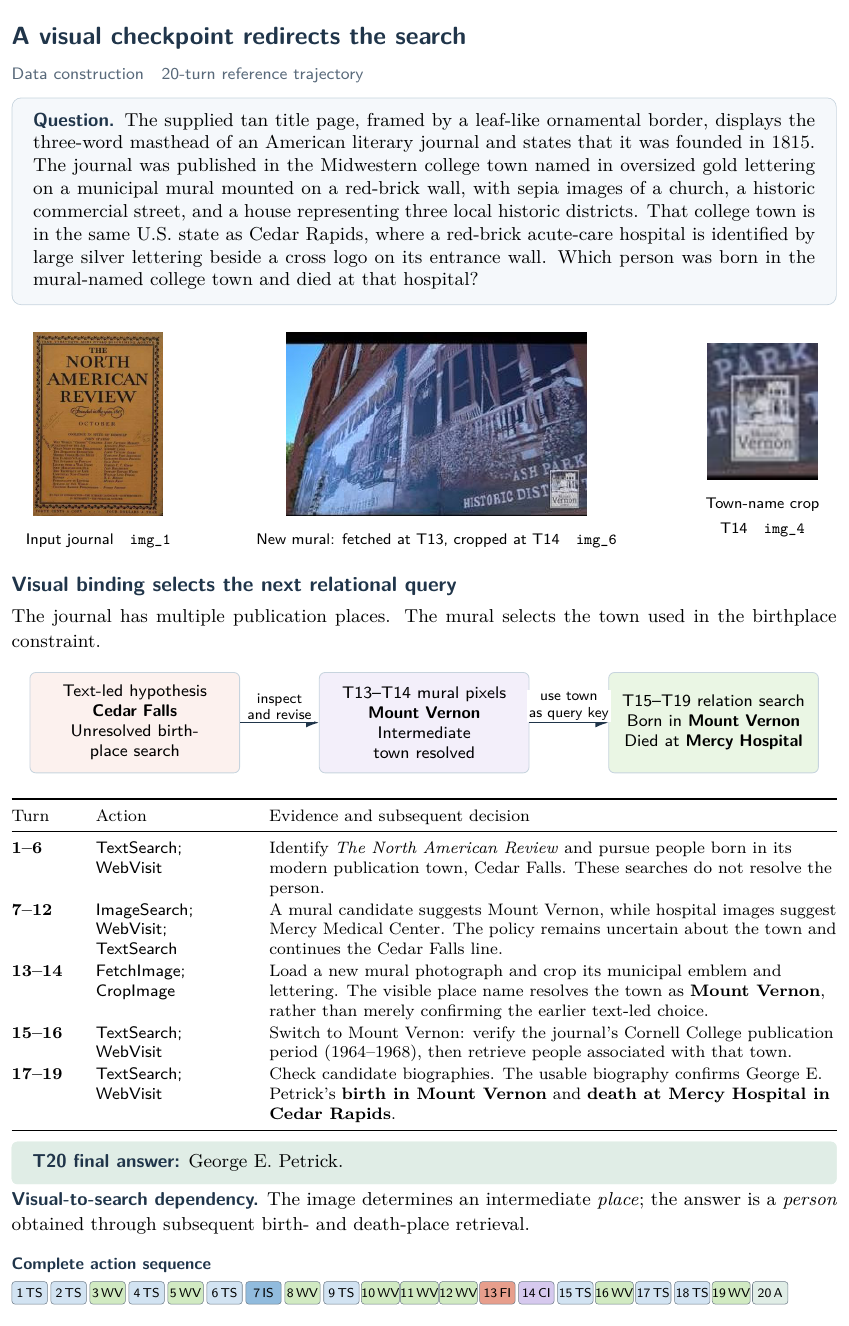}
  \caption{\textbf{Intermediate visual disambiguation in a constructed question.} The policy initially follows Cedar Falls, a publication place of the journal. A newly retrieved mural and its town-name crop resolve Mount Vernon instead, redirecting the subsequent birthplace and hospital-death searches that identify George E.\ Petrick.}
  \label{fig:neos_visualization_constructed_album}
\end{figure}

\begin{figure}[!htbp]
  \centering
  \includegraphics[width=\linewidth,height=0.84\textheight,keepaspectratio]{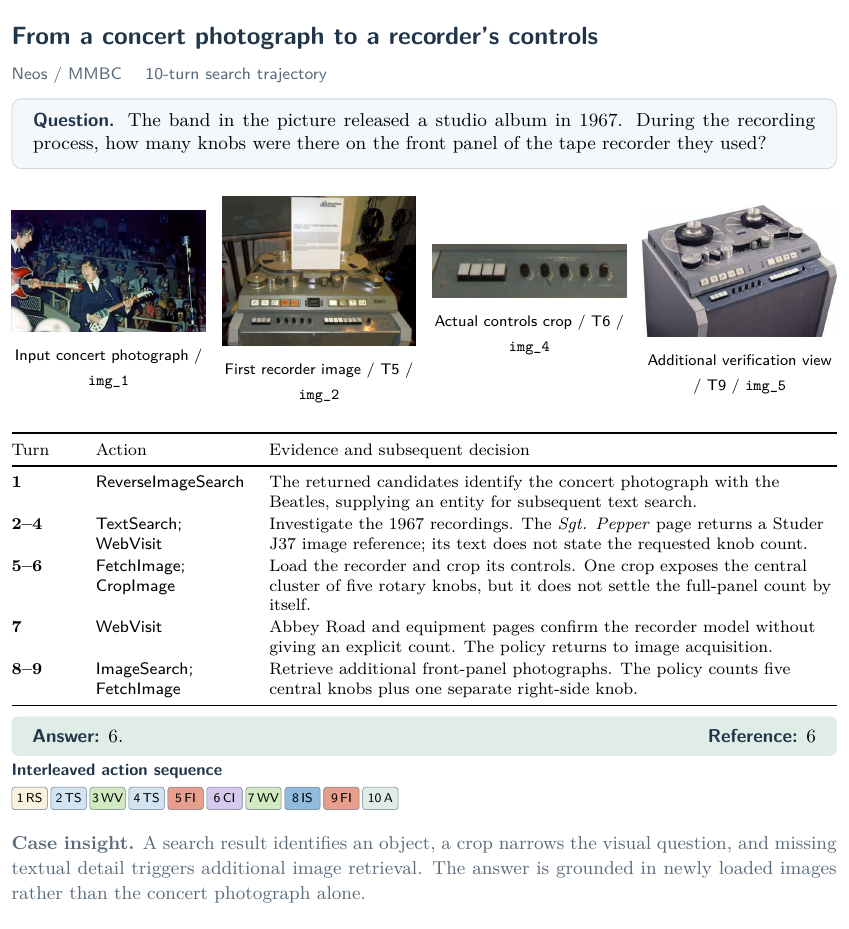}
  \caption{\textbf{MMBC example}: a concert photograph leads to a recorder model. A partial crop is followed by additional image acquisition to verify the complete front-panel knob count.}
  \label{fig:neos_visualization_mmbc}
\end{figure}

\begin{figure}[!htbp]
  \centering
  \includegraphics[width=\linewidth,height=0.84\textheight,keepaspectratio]{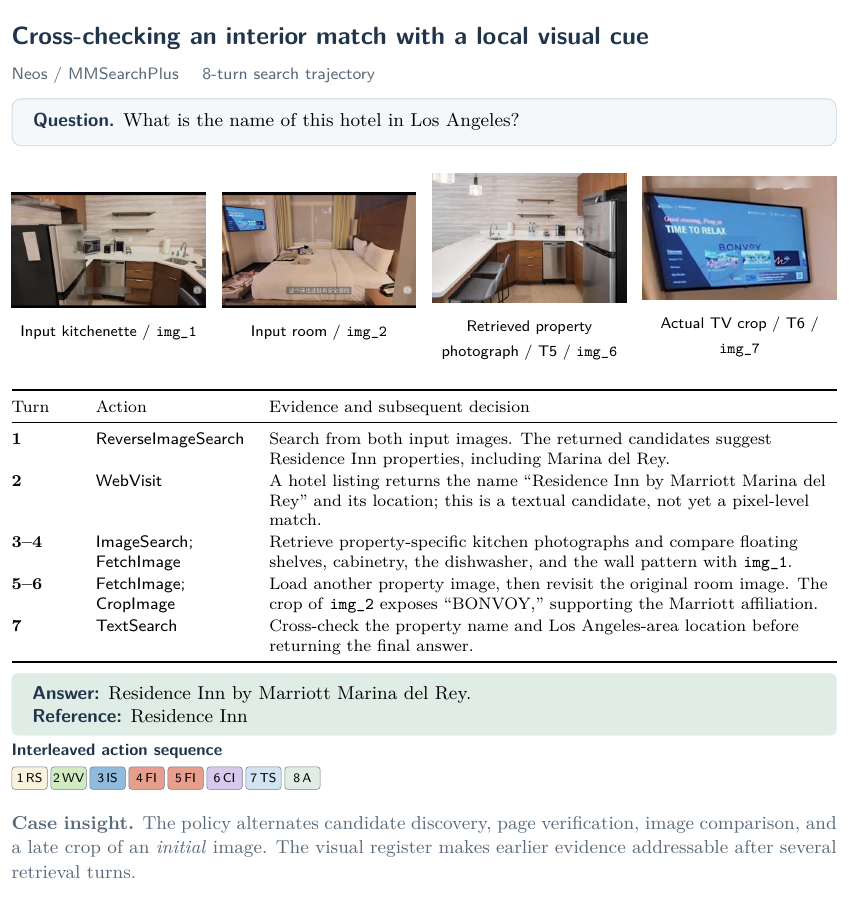}
  \caption{\textbf{MMSearchPlus example}: candidate hotel interiors are checked against retrieved photographs and source pages. A late crop of an initial image exposes the Bonvoy cue, allowing earlier visual evidence to be revisited during search.}
  \label{fig:neos_visualization_mmsearchplus}
\end{figure}

\begin{figure}[!htbp]
  \centering
  \includegraphics[width=\linewidth,height=0.84\textheight,keepaspectratio]{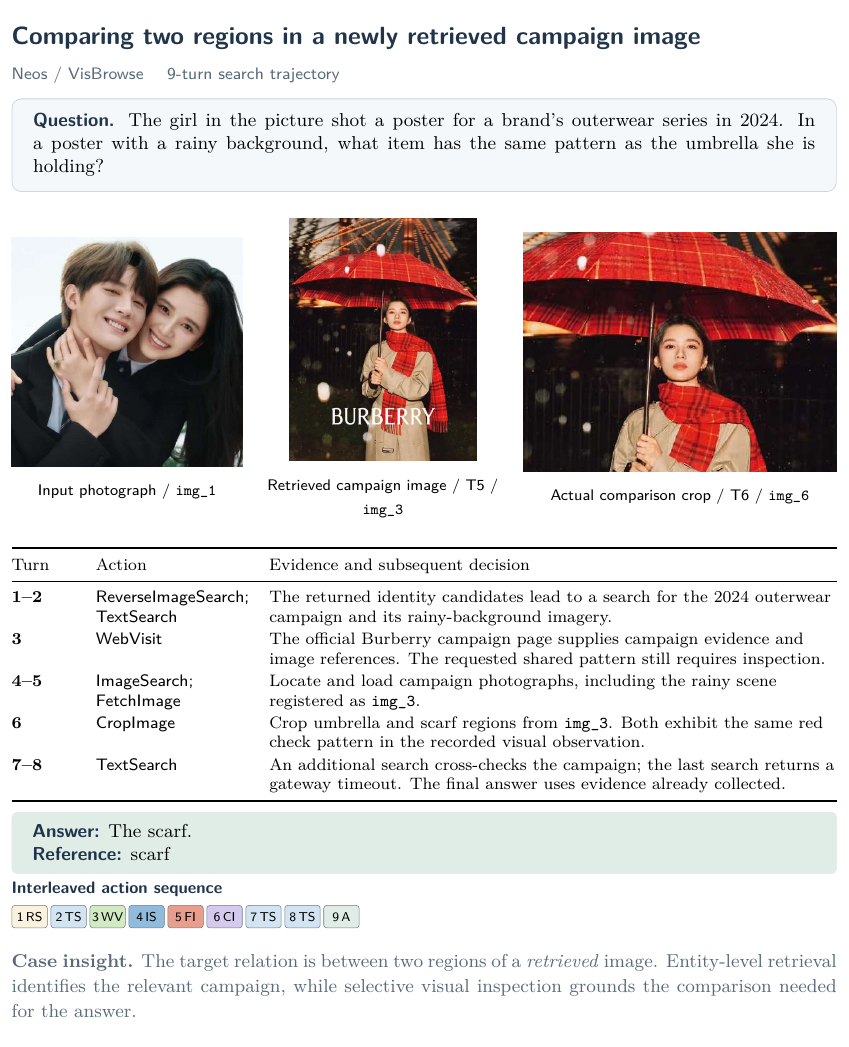}
  \caption{\textbf{VisBrowse example}: campaign retrieval is followed by fetching and cropping a new image to compare the umbrella and scarf, connecting entity-level search with region-level visual reasoning.}
  \label{fig:neos_visualization_visbrowse}
\end{figure}
\begin{figure}[!htbp]
  \centering
  \includegraphics[width=\linewidth,height=0.84\textheight,keepaspectratio]{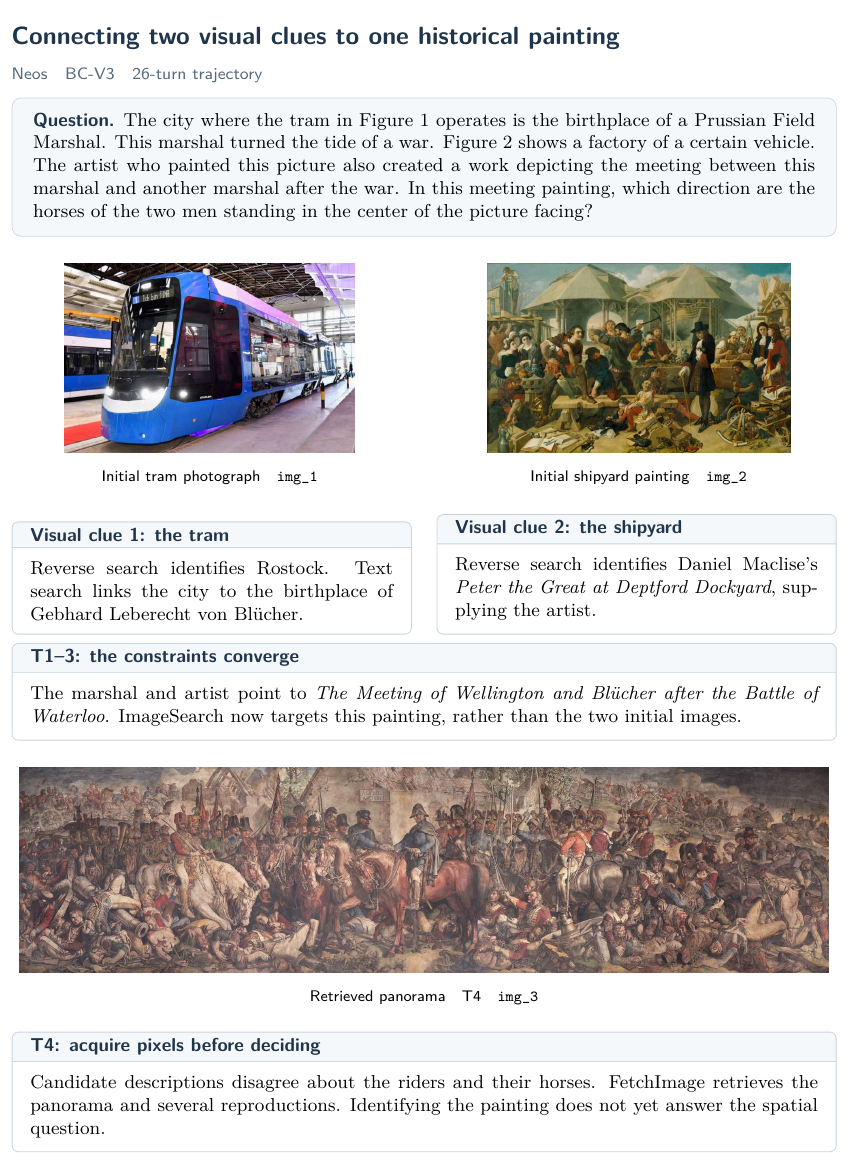}
  \caption{\textbf{Long-horizon BC-V3 example}, part one. Two initial images provide different entity constraints: a tram leads to a city and a marshal, while a shipyard painting identifies an artist. Their intersection determines a new painting to retrieve and inspect.}
  \label{fig:neos_visualization_history_identification}
\end{figure}

\begin{figure}[!htbp]
  \centering
  \includegraphics[width=\linewidth,height=0.84\textheight,keepaspectratio]{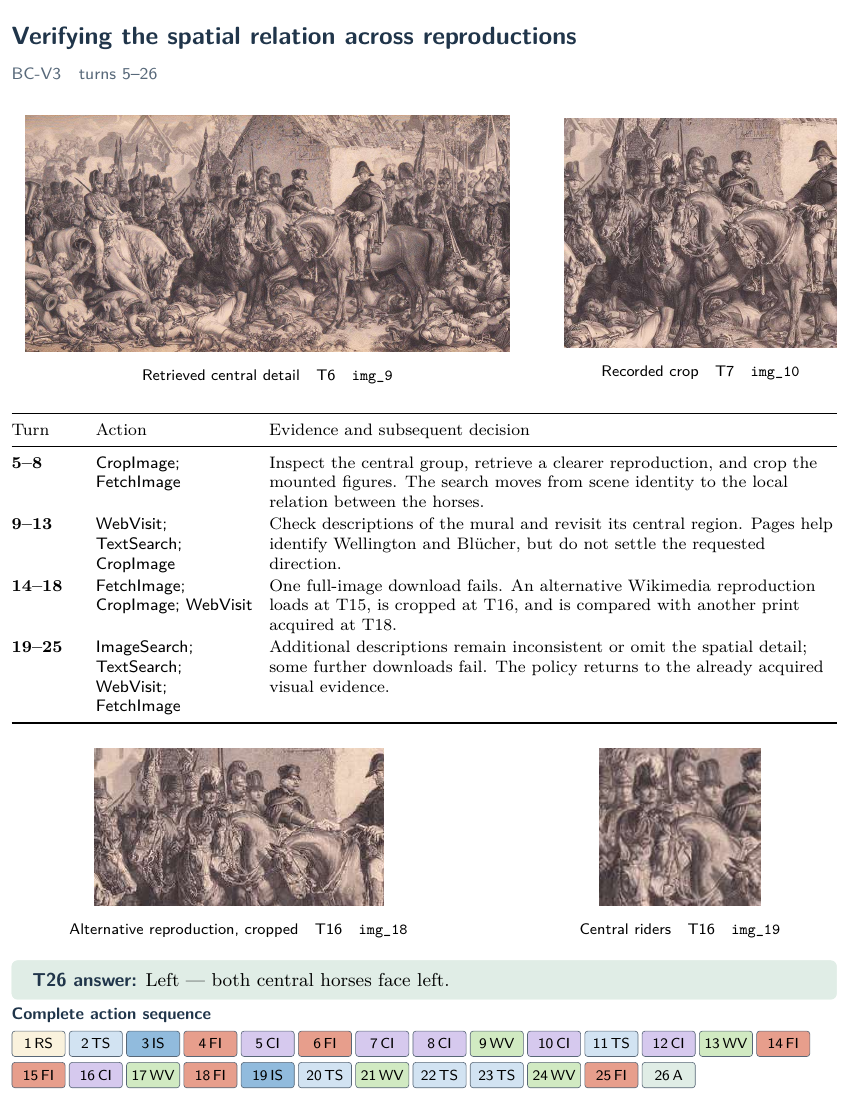}
  \caption{\textbf{Long-horizon BC-V3 example}, part two. The 26-turn trajectory alternates between image acquisition, local crops, and source-page checks. Conflicting descriptions and unsuccessful downloads lead the policy back to previously acquired pixels to answer the spatial question.}
  \label{fig:neos_visualization_history_verification}
\end{figure}

\begin{figure}[!htbp]
  \centering
  \includegraphics[width=\linewidth,height=0.84\textheight,keepaspectratio]{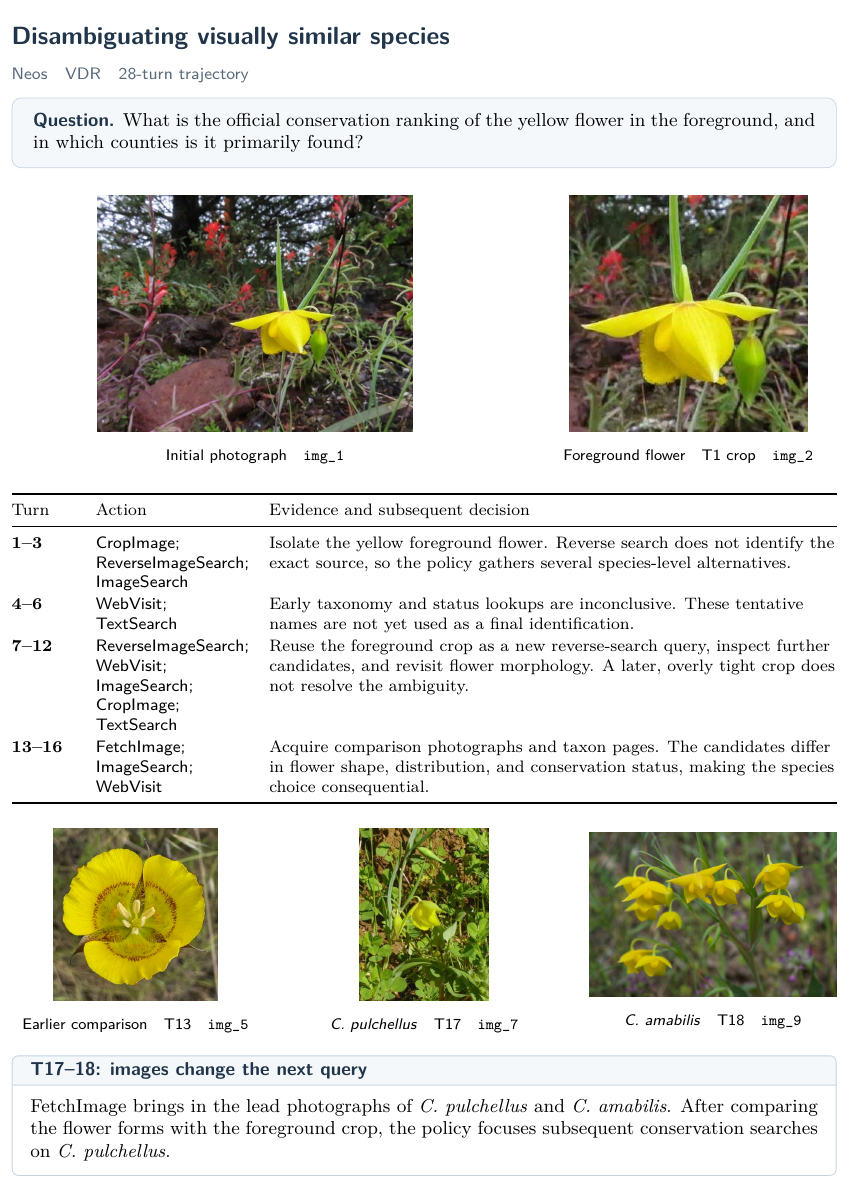}
  \caption{\textbf{Long-horizon VDR example}, part one. A foreground crop is reused for reverse search, and comparison photographs help distinguish visually similar species before the policy commits to species-specific factual retrieval.}
  \label{fig:neos_visualization_flower_identification}
\end{figure}

\begin{figure}[!htbp]
  \centering
  \includegraphics[width=\linewidth,height=0.84\textheight,keepaspectratio]{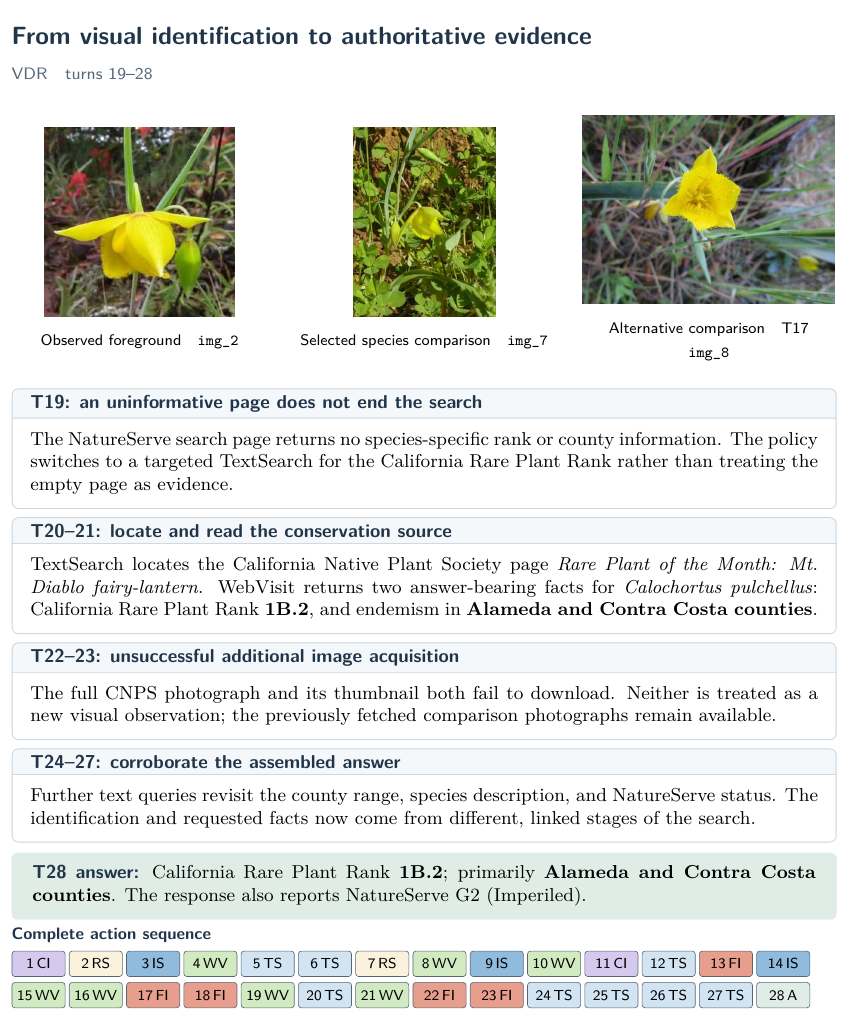}
  \caption{\textbf{Long-horizon VDR example}, part two. The 28-turn trajectory connects visual identification with conservation evidence. An uninformative database page triggers another search, after which a conservation organization supplies the rank and county range.}
  \label{fig:neos_visualization_flower_facts}
\end{figure}

\clearpage

\end{document}